\documentclass[12pt,preprint]{elsarticle}

\usepackage{lineno, hyperref, amssymb, graphicx, float, subcaption}
\usepackage{amsmath,bm, amsthm, mathtools, tabularx, xcolor}
\usepackage[ruled,vlined]{algorithm2e}
\SetKwComment{Comment}{$\triangleright$\ }{}
\usepackage{color}
\usepackage{geometry}
\usepackage{pdflscape}
\usepackage{threeparttable} % Foot notes under tables
\modulolinenumbers[5]

\newtheorem{theorem}{Theorem}

\newtheorem*{remark}{Remark}

\newcommand{\norm}[1]{\left\lVert#1\right\rVert}

\journal{Neural Networks}

\title{Transformers for Modeling Physical Systems}
\date{\today}

\author[label1]{Nicholas Geneva}
\ead{ngeneva@nd.edu}

\author[label1]{Nicholas Zabaras\corref{cor1}}
\ead{nzabaras@gmail.com}
\ead[url]{https://cics.nd.edu/}

\address[label1]{Scientific Computing and Artificial Intelligence (SCAI) Laboratory, University of Notre Dame, 311 Cushing Hall, Notre Dame, IN 46556, USA}
\cortext[cor1]{Corresponding author}

\begin{document}

\begin{abstract}
    Transformers are widely used in natural language processing due to their ability to model longer-term dependencies in text.
    Although these models achieve state-of-the-art performance for many language related tasks, their applicability outside of the natural language processing field has been minimal.
    In this work, we propose the use of transformer models for the prediction of dynamical systems representative of physical phenomena.
    The use of Koopman based embeddings provide a unique and powerful method for projecting any dynamical system into a vector representation which can then be predicted by a transformer.
    The proposed model is able to accurately predict various dynamical systems and outperform classical methods that are commonly used in the scientific machine learning literature.\footnote{Code available at: \href{https://github.com/zabaras/transformer-physx}{https://github.com/zabaras/transformer-physx}.}
\end{abstract}


\maketitle
 
% \linenumbers

\section{Introduction}
\noindent
The transformer model~\citep{vaswani2017attention}, built on self-attention, has largely become the state-of-the-art approach for a large set of natural language processing (NLP) tasks including language modeling, text classification, question answering, etc.
Although more recent transformer work is focused on unsupervised pre-training of extremely large models~\citep{devlin2018bert, radford2019language, dai2019transformer, liu2019roberta}, the original transformer model garnered attention due to its ability to out-perform other state-of-the-art methods by learning longer-term dependencies without recurrent connections.
Given that the transformer model was originally developed for NLP, nearly all related work has been rightfully confined within this field with only a few exceptions.
Here, we focus on the development of transformers to model dynamical systems that can replace otherwise expensive numerical solvers.
In other words, we are interested in using transformers to learn the language of physics.

The surrogate modeling of physical systems is a research field that has existed for several decades and is a large ongoing effort in scientific machine learning.
Past literature has explored multiple surrogate approaches including Gaussian processes~\cite{bilionis2013multi, bilionis2016bayesian, atkinson2019structured}, polynomial chaos expansions~\cite{xiu2002wiener}, reduced-order models~\cite{chakraborty2018efficient, gao2020non}, reservoir computing~\cite{tanaka2019recent} and deep neural networks~\cite{zhu2018bayesian, tripathy2018deep, geneva2020modeling}.
A surrogate model is defined as a computationally inexpensive approximate model of a physical phenomenon that is designed to replace an expensive computational solver that would otherwise be needed to resolve the system of interest.
An important characteristic of surrogate models is their ability to model a distribution of initial or boundary conditions rather than learning just one solution.
This is arguably essential for the justification of training a deep learning model versus using a standard numerical solver, particularly in the context of utilizing deep learning methods which tend to have expensive training procedures.
The most tangible applications of surrogates are for optimization, design and inverse problems where many repeated simulations are typically needed.

Standard deep neural network architectures such as auto-regressive~\citep{mo2019deep, geneva2020modeling}, residual/Euler~\citep{gonzalez1998identification, sanchezgonzalez2020learning}, recurrent and LSTM based models~\citep{mo2019deep, tang2020deep, maulik2020recurrent, geneva2020multi} have been largely demonstrated to be effective at modeling various physical dynamics.
Such models generally rely on the most recent time-steps to provide complete information on the current and past state of the system's evolution.
Approaches that meld numerical time--integration methods with neural networks have also proven to be fairly successful, e.g.~\cite{wang1998runge, zhu2018convolutional, wessels2020neural}, but  have a fixed temporal window from which information is provided.
Present machine learning models lack generalizable time cognizant capabilities to predict multi-time-scale phenomena present in systems including turbulent fluid flow, multi-scale materials modeling, molecular dynamics, chemical processes, etc.
Much work is needed to scale such deep learning models to complex physical systems that are of scientific and industrial interest.
This work deviates from this pre-existing literature by investigating the use of transformers for the prediction of physical systems, relying entirely on self-attention to surrogate model dynamics.
In the recent work of~\cite{shalova2020tensorized}, such self-attention models were tested to learn single solutions of several low-dimensional ordinary differential equations.

The novel contributions of this paper are as follows:
(a) The application of self-attention transformer models for modeling physical dynamics;
(b) The use of Koopman dynamics for developing  physics inspired embeddings of high-dimensional systems with connections to embedding methods seen in NLP;
(c) Discussion of the relations between self-attention with traditional numerical time-integration;
(d) Demonstration of our model on high-dimensional partial differential equation problems that include  chaotic dynamics, fluid flows and reaction-diffusion systems.
To the authors best knowledge, this is the first work to explore transformer NLP architectures for the surrogate modeling of physical systems.
The remainder of this paper is as follows:
In Section~\ref{sec:methodology}, the machine learning methodology is discussed including the transformer decoder model in Section~\ref{sec:transformer} and the Koopman embedding model in Section~\ref{sec:embedding}.
Following in Section~\ref{sec:numerical}, the proposed model is implemented for a set of numerical examples of different dynamical nature.
This includes classical chaotic dynamics in Section~\ref{sec:lorenz}, periodic fluid dynamics in Section~\ref{sec:cylinder} and three-dimensional reaction-diffusion dynamics in Section~\ref{sec:gray_scott}.
Lastly, concluding discussion and future directions are given in Section~\ref{sec:conclusion}.

\section{Methods}
\label{sec:methodology}
\noindent
We are interested in systems that can be described through a dynamical ordinary or partial differential equation of the form:
\begin{equation}
    \begin{gathered}
        \bm{\phi}_{t} = F\left(\bm{x},\bm{\phi}(t,\bm{x},\bm{\eta}), \nabla_{\bm{x}}\bm{\phi}, \nabla_{\bm{x}}^{2}\bm{\phi},  \bm{\phi}\cdot\nabla_{\bm{x}}\bm{\phi}, \ldots \right),\\
        \quad t \in \mathcal{T} \subset \mathbb{R}^{+}, \quad \bm{x} \in \Omega \subset \mathbb{R}^{m},
    \end{gathered}
    \label{eq:pde}
\end{equation}
in which  $\bm{\phi} \in \mathbb{R}^{n}$ is the solution of this differential equation of $n$ state variables with parameters $\bm{\eta}$,  in the time interval $\mathcal{T}$ and spatial domain $\Omega$ with a boundary $\Gamma \subset \Omega$.
This general form can embody a vast spectrum of physical phenomena including fluid flow and transport processes, mechanics and materials physics, and molecular dynamics.
In this work, we are interested in learning the set of solutions for a distribution of initial conditions $\bm{\phi}_{0} \sim p(\bm{\phi}_{0})$, boundary conditions $\mathcal{B}(\bm{\phi}) \sim p\left(\mathcal{B}\right) \forall \bm{x}\in \Gamma$ or equation parameters $\bm{\eta}\sim p\left(\bm{\eta}\right)$.
This accounts for modeling initial value, boundary value and stochastic problems.

\begin{figure}[h]
    \centering
    \includegraphics[width=\textwidth]{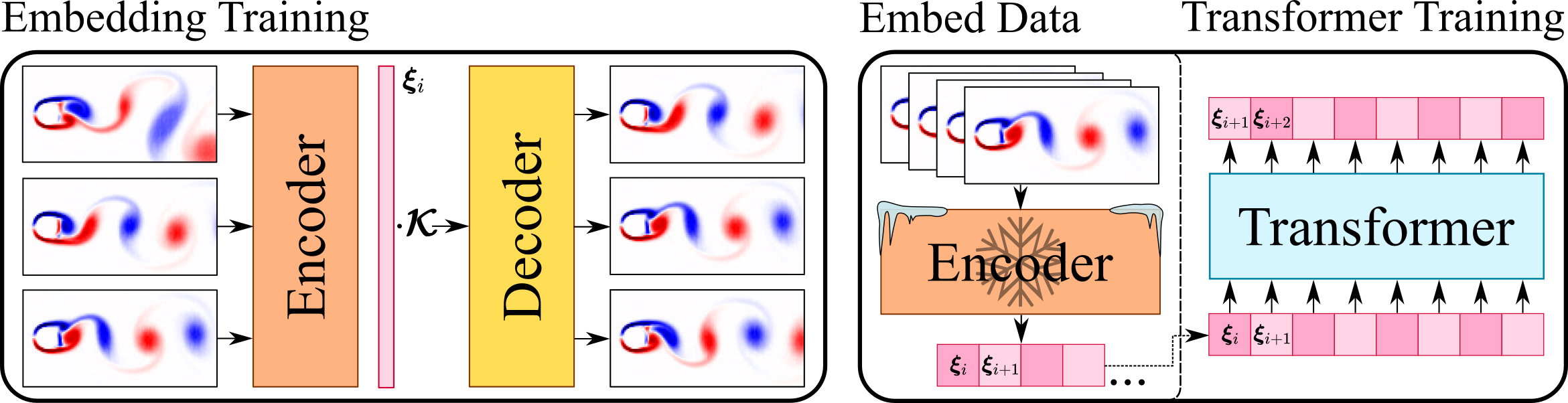}
    \caption{The two training stages for modeling physical dynamics using transformers. (Left to right) The embedding model is first trained using Koopman based dynamics. The embedding model is then frozen (fixed), all training data is embedded and the transformer is trained in the embedded space.}
    \label{fig:training}
\end{figure}

To make the modeling of such dynamical systems applicable to the use of modern machine learning architectures, the continuous solution is discretized in both the spatial and temporal domains such that the solution of the differential equation is $\Phi = \left(\bm{\phi}_{0}, \bm{\phi}_{1},\ldots \bm{\phi}_{T}\right); \bm{\phi}_{i} \in \mathbb{R}^{n\times d}$, for which $\bm{\phi}_{i}$ has been discretized by $d$ points in $\Omega$. 
We assume an initial state  $\bm{\phi}_{0}$ and that the time interval $\mathcal{T}$ is discretized by $T$ time-steps with a time-step size $\Delta t$.
Hence, we pose the modeling a dynamical system as a time-series problem.
The proposed machine learning methodology has two core components: the transformer for modeling dynamics and the embedding network for projecting physical states into a vector representation.
Similar to NLP, the embedding model is trained prior to the transformer.
This embedding model is then frozen and the entire data-set is converted to the embedded space in which the transformer is then trained as illustrated in Fig.~\ref{fig:training}.
During testing, the embedding decoder is used to reconstruct the physical states from the transformer's predictions.

\subsection{Transformer}
\label{sec:transformer}
\noindent
The transformer model was originally designed with NLP as the sole application with word vector embeddings of a passage of text being the primary input~\citep{vaswani2017attention}.
However, recent works have explored using attention mechanisms for different machine learning tasks~\citep{velivckovic2017graph, zhang2019self, fu2019dual} and a few investigate the use of transformers for applications outside of the NLP field~\citep{chen2020generative}.
This suggests that self-attention and in particular transformer models may work well for any problem that can be posed as a sequence of vectors.

\subsubsection{Transformer Decoder}
\noindent
In this work, the primary input to the transformer will be an embedded dynamical system, $\Xi = \left(\bm{\xi}_{0}, \bm{\xi}_{1}, \ldots. \bm{\xi}_{T}\right)$, where the embedded state at time-step $i$ is denoted as $\bm{\xi}_{i}\in\mathbb{R}^{e}$.
Given that we are interested in the prediction of a physical time series, this motivates the usage of a language modeling architecture that is designed for the sequential prediction of words in a body of text.
We select the transformer decoder architecture used in the Generative Pre-trained Transformer (GPT) models~\citep{radford2018improving, radford2019language}.
\begin{figure}[h]
    \centering
    \includegraphics[width=\textwidth]{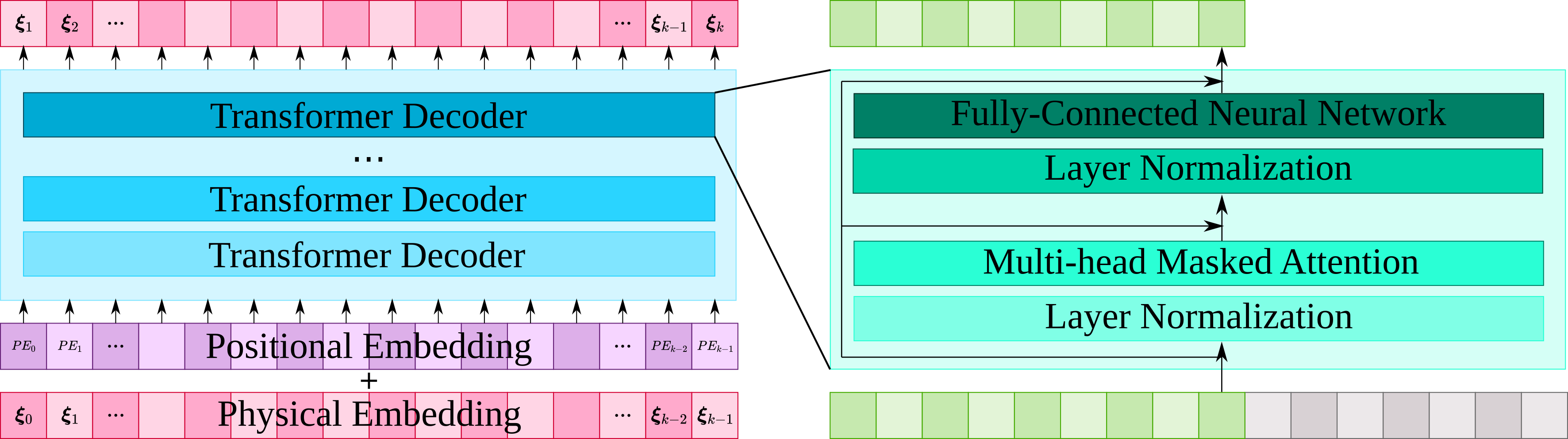}
    \caption{The transformer decoder model used for the prediction of physical dynamics.}
    \label{fig:transformer}
\end{figure}
Our model follows the GPT-2 architecture based on the implementation in the Hugging Face transformer repository~\citep{wolf2019HuggingFacesTS}, but is significantly smaller in size than these modern NLP transformers.
This model consists of a stack of transformer decoder layers that use masked attention, as depicted in Fig.~\ref{fig:transformer}.
The input to the transformer is the embedded representation of the physical system from the embedding model and a sinusoidal positional encoding proposed in the original transformer~\citep{vaswani2017attention}.
Since the transformer does not contain any recurrent or convolutional operation, information regarding the relative position of the embedded input sequence must be provided.
The positional embedding is defined by the following sine and cosine functions:
\begin{equation}
    PE_{pos,2j} = \textrm{sin}\left(pos/10000^{2j/e}\right), \quad PE_{pos,2j+1} = \textrm{cos}\left(pos/10000^{2j/e}\right),
\end{equation}
for the $2j$ and $2j+1$ elements in the embedded vector, $\bm{\xi}_{i}$. 
$pos$ is the embedded vector's relative or global position in the input time series.
To train the model, consider a data set of $D$ embedded i.i.d. time-series $\mathcal{D}=\left\{\Xi^{i}\right\}^{D}_{i=1}$ for which we can use the standard time-series Markov model (language modeling) log-likelihood:
\begin{equation}
    L_{\mathcal{D}} = \sum_{i}^{D}\sum_{j}^{T} -\log p\left(\bm{\xi}^{i}_{j}| \bm{\xi}^{i}_{j-k}, \ldots, \bm{\xi}^{i}_{j-1}, \bm{\theta}\right),
\end{equation}
where $\bm{\theta}$ are the model's parameters and $k$ is the transformer's context window.
Contrary to the standard NLP approach which poses the likelihood as a softmax over a dictionary of tokens, the likelihood here is taken as a standard Gaussian between the transformer's prediction and the target embedded value resulting in a $L_2$ loss.
This is due to the fact that the solution to most physical systems cannot be condensed to a discrete finite set.
Thus tokenization into a finite dictionary not possible and a softmax approach not applicable.
Training is the standard auto-regressive method used in GPT~\citep{radford2018improving}, as opposed to the word masking~\citep{devlin2018bert}, constrained to the embedded space.
The physical states, $\bm{\phi}_{i}$, have the potential to be very high-dimensional and training the transformer in the lower-dimensional embedded space can significantly lower training costs. 

\subsubsection{Self-Attention}
\noindent
Self-attention is the main mechanism that allows the transformer to learn temporal dependencies.
Prior to the seminal transformer paper, several works had already proposed a handful of different attention models typically integrated into recurrent models for language processing~\cite{graves2014neural, bahdanau2014neural, luong2015effective}.
With the popularization of the transformer model~\cite{vaswani2017attention}, the most commonly used attention model is the scaled-dot product attention:
\begin{gather}
    \bm{k}_{i} = \mathcal{F}_{k}(\bm{x}_{i}), \quad  \bm{q}_{i} = \mathcal{F}_{q}(\bm{x}_{i}), \quad  \bm{v}_{i} = \mathcal{F}_{v}(\bm{x}_{i}) \\
    \bm{c}_{n} = \sum_{i=1}^{k} \alpha_{n,i} \bm{v}_{i}, \qquad \alpha_{n,i} = \frac{exp(\bm{q}^{T}_{n} \bm{k}_{i} / \sqrt{d_{k}})}{\sum_{j=1}^{k} exp(\bm{q}^{T}_{n} \bm{k}_{j} / \sqrt{d_{k}})},
\end{gather}
in which we use $\bm{x}_{i}\in \mathbb{R}^{d}$ and $ \bm{c}_{i}\in\mathbb{R}^{d_{v}}$ to denote an arbitrary input and context output, respectively.
$\bm{k} \in \mathbb{R}^{d_{k}}$, $\bm{q} \in \mathbb{R}^{d_{k}}$ and $\bm{v} \in \mathbb{R}^{d_{v}}$ are referred to as the key, query, and value vectors, respectively, calculated using neural networks $\mathcal{F}_{k}$, $\mathcal{F}_{q}$ and $\mathcal{F}_{v}$.
The attention score, $\alpha_{n,i}$, is calculated by the soft-max of the dot product between the query and key vectors scaled by the dimension $d_{k}$.
Due to the soft-max calculation note that the attention scores always sum to one, $\sum_{i=1}^{k}\alpha_{n,i}=1$, for every input.

The attention calculation can be condensed for the entire context length of the model by a matrix representation $\bm{C} = softmax\left(\bm{Q}\bm{K}^{T}/\sqrt{d_{k}}\right)\bm{V}$, where $\bm{Q}\in\mathbb{R}^{k \times d_{k}}$, $\bm{K}\in\mathbb{R}^{k \times d_{k}}$ and $\bm{V}\in\mathbb{R}^{k \times d_{v}}$.
As illustrated in Fig.~\ref{fig:transformer}, self-attention is typically implemented with a residual connection with multiple independent attention calculations referred to as attention heads~\cite{vaswani2017attention}.
While computationally inexpensive to evaluate, the memory requirement of scaled-dot product attention can become increasingly cumbersome as the context length, $k$, increases.
Hence, several methods for approximating this calculation have been proposed which include the use of kernels and random projections in an attempt to lower the dimensionality of the self-attention calculation without loss of predictive accuracy~\cite{sukhbaatar2019adaptive, sukhbaatar2019augmenting, kitaev2020reformer}.

While designed for natural language processing, in the context of dynamical systems, self-attention bears a very similar form to numerical time-integration methods.
The use of time-integration methods such as Runge-Kutta schemes or Euler methods can be found in Neural ODEs~\cite{chen2018neural, dupont2019augmented} and Res-Net based models~\cite{gonzalez1998identification, sanchezgonzalez2020learning}.
Self-attention offers a much larger learning capacity than using these traditional numerical methods for learning a unique latent time-integration method.
The function space of a self-attention layer with a residual connection contains the space of explicit linear time-integration methods within an arbitrarily small non-zero amount of error.

\begin{theorem}
    Consider a dynamical system of the form, $d\bm{\phi}/dt=f(t,\bm{\phi}), \ \bm{\phi}(t)\in\mathbb{R}^{d}$, where $f:\mathbb{R}\times\mathbb{R}^{d}\rightarrow \mathbb{R}^{d}$ is Lipschitz continuous with respect to $\bm{\phi}$ and continuous in $t$.
    Let the function $\mathcal{A}_{\theta}\left(t,\bm{\phi}_{t-k\Delta t:t}\right) =  \hat{\bm{\phi}}_{t+\Delta t}$ be a self-attention layer with a residual connection, of context length $k$ and containing learnable parameters $\bm{\theta}\in\mathbb{R}^{d_{\theta}}$.
    Suppose $A:=\left\{\mathcal{A}_{\theta}\left(t,\bm{\phi}\right)\, |\, \forall \bm{\theta} \in \mathbb{R}^{d_{\theta}}\right\}$ be the set of all possible self-attention calculations.
    Let $\mathcal{M}_{i}\left(t,\bm{\phi}_{t-(i-1)\Delta t:t}\right) =  \bm{\phi}_{t+\Delta t}$ be the  $i^{\textrm{th}}$ order explicit Adams method time-integrator. 
    The set of functions up to $k^{th}$ order $M := \left\{\mathcal{M}_{i}\, |\, 1\le i \le k \right\}$ is a subset of $A$ such that $\exists\, \mathcal{A}_{\theta_{i}} \in A$ s.t. $||\mathcal{M}_{i} - \mathcal{A}_{\theta_{i}}||_{\infty} < O(\epsilon)$ for any $\mathcal{M}_{i} \in M$ and $\epsilon > 0$.
\end{theorem}
\noindent \textit{Proof}. The state variables are discretized w.r.t. time such that $\bm{\phi}_{i}=\bm{\phi}(i\Delta t);\, t_{i}=i \Delta t$. 
The general definition for linear multi-step methods follows:
\begin{multline}
    \bm{\phi}_{n+s} + a_{s-1}\cdot \bm{\phi}_{n+s-1} + ... + a_{0}\cdot \bm{\phi}_{n} \\
    = \Delta t \cdot \left(b_{s}\cdot f\left(t_{n+s},\bm{\phi}_{n+s}\right) + b_{s-1}\cdot f\left(t_{n+s-1},\bm{\phi}_{n+s-1}\right) + ... + b_{0}\cdot f\left(t_{n},\bm{\phi}_{n}\right)\right),
\end{multline}
in which $a_{i}$ and $b_{i}$ are coefficients determined by the integration method.
For explicit Adams methods, the $\bm{\phi}_{n+s}$ is directly computed with $b_{s}=0$, $a_{s-1}=-1$ and $a_{0:s-2}=0$.
The generalized formula for $s$-step explicit Adams methods can be represented as the following linear combination:
\begin{equation}
    \mathcal{M}_{s} = \bm{\phi}_{n+s} = \bm{\phi}_{n+s-1} + \Delta t \sum_{j=0}^{s-1}b_{j}f\left(t_{n+j},\bm{\phi}_{n+j}\right),
    \label{eq:multi-step-sum}
\end{equation}
which encapsulates up to and including $s^{th}$ order time integration~\cite{stoer2013introduction}.
Consider a residual scaled dot-product self-attention calculation with context length, $s$, for output prediction $\hat{\bm{\phi}}_{n+s}$, input states $\bm{\phi}_{n:n+s-1}$ and time $t$:
\begin{equation}
    \mathcal{A}_{\theta_{i}} = \hat{\bm{\phi}}_{n+s} = \bm{\phi}_{n+s-1} + \sum_{i=0}^{s-1}\alpha_{n+s-1,i}\, \bm{v}_{i}, \quad \alpha_{n+s-1,i} = \frac{exp(\bm{q}^{T}_{n+s-1} \bm{k}_{i} / \sqrt{d_{k}})}{\sum_{j=0}^{s-1} exp(\bm{q}^{T}_{n+s-1} \bm{k}_{j} / \sqrt{d_{k}})},
    \label{eq:attention-sum}
\end{equation}
for which $\bm{k}_{i} = \mathcal{F}_{k}(t_{n+i},\bm{\phi}_{n+i})$, $\bm{q}_{i}= \mathcal{F}_{q}(t_{n+i},\bm{\phi}_{n+i})$ and $\bm{v}_{i}= \mathcal{F}_{v}(t_{n+i},\bm{\phi}_{n+i})$ vectors are outputs of differentiable functions parameterized by neural networks:
\begin{equation}
    \mathcal{F}_{k}:\mathbb{R}\times\mathbb{R}^{d} \rightarrow \mathbb{R}^{d_{k}}, \quad \mathcal{F}_{q}:\mathbb{R}\times\mathbb{R}^{d} \rightarrow \mathbb{R}^{d_{k}}, \quad \mathcal{F}_{v}:\mathbb{R}\times\mathbb{R}^{d} \rightarrow \mathbb{R}^{d_{v}}.
\end{equation}
It suffices to show that there exists a form of Eq.~(\ref{eq:attention-sum}) that is equal to Eq.~(\ref{eq:multi-step-sum}) within a error order $O(\epsilon)$ such that 
\begin{equation}
\left|\left|\bm{\phi}_{n+s} - \hat{\bm{\phi}}_{n+s}\right|\right|_{\infty} < O(\epsilon). \label{eq:error}
\end{equation}
To this end, although $\mathcal{F}_q$ and $\mathcal{F}_k$ are fully-trainable neural networks, herein we consider a fixed single-entry query and key vectors resulting in the following simplified self-attention scores:
\begin{equation}
\begin{gathered}
    \bm{q}_{i} = \mathcal{F}_{q}(t_{n+i},\bm{\phi}_{n+i}) = \sqrt{d_{k}}\, \bm{e}_{m}, \quad \bm{k}_{i} = \mathcal{F}_{k}(t_{n+i},\bm{\phi}_{n+i}) = \log(b_{i})\, \bm{e}_{m}, \\
    \alpha_{n+s-1,i} = \frac{exp(\log(b_{i}))}{\sum_{j=0}^{s-1} exp(\log(b_{j}))} = \frac{b_{i}}{\sum_{j=0}^{s-1} b_{j}},
\end{gathered}   
\label{eq:attention}
\end{equation}
where $\bm{e}_{m}\in \mathbb{R}^{d_{k}}$ is a unit vector
with all elements set to zero except the  $m$-th element that is set to $1$.
By the universal approximation theorem~\cite{hornik1989multilayer, cybenko1989approximation, hornik1991approximation}, the neural network, $\mathcal{F}_{v}$, is assumed to be of sufficient capacity such that it can approximate the r.h.s. of the governing equation:
\begin{equation}
    \begin{gathered}
    \bm{v}_{i} = \mathcal{F}_{v}(t_{n+i},\bm{\phi}_{n+i}) = c\,  f(t_{n+i},\bm{\phi}_{n+i}) + O(\epsilon), \quad \epsilon > 0,
    \end{gathered}
    \label{eq:key}
\end{equation}
where the constant $c$ is taken as $c = \Delta t \sum_{j=0}^{s-1} b_{j}$. 
Equations~(\ref{eq:attention}) and~(\ref{eq:key}) can then be combined leading to the following:

\begin{equation}
    \begin{aligned}
        \bm{\phi}_{n+s-1} + \sum_{i=0}^{s-1} \alpha_{n+s-1,i}\,  \bm{v}_{i} &=    \bm{\phi}_{n+s-1} + \sum_{i=0}^{s-1} \frac{b_{i}}{\sum_{j=0}^{s-1} b_{j}}\,  c\,  f(t_{n+i},\bm{\phi}_{n+i}) + O(\epsilon),\\
         &= \bm{\phi}_{n+s-1} + \Delta t \sum_{i=0}^{s-1} b_{i} f(t_{n+i},\bm{\phi}_{n+i}) + O(\epsilon).
        \label{eq:atten-eq-multi}
    \end{aligned}
\end{equation}
For a large capacity neural network, we can assume that the error can be made arbitrarily small  ($\epsilon \rightarrow 0$) and Eq.~(\ref{eq:error}) is proved. 
Therefore, the form of the explicit Adams multi-step methods of order $\leq s$ can be captured by a residual self-attention transformer layer. \qed

A single transformer layer is significantly more expressive than any numerical time integration scheme, enabling it to learn more complex temporal dependencies.
However, the linear combination of weighted features from past time-steps is a commonality between self-attention and numerical time-integration methods.
This mathematical similarity indicates self-attention models may be better suited for learning dynamical systems than alternative deep learning approaches.
In particular traditional Euler time integration present in various models for modeling dynamics~\cite{gonzalez1998identification, sanchezgonzalez2020learning, lu2018beyond} is a specific case of this single layer attention calculation.

\begin{remark}
The inclusion of time, $t$, as an input is required for non-autonomous systems with explicit time-dependent terms.
This has surprising connections to the implementation of the transformer which uses a positional embedding.
Although the positional embedding is included for the sake of denoting the order of the input to the transformer, using current time of the given input state can accomplish the same goal.
\end{remark}

\subsection{Embedding Model}
\label{sec:embedding}
\noindent
The second major component of the machine learning methodology is the embedding model responsible for projecting the discretized physical state space into a 1D vector representation.
In NLP, the standard approach is to tokenize then embed a finite vocabulary of words, syllables or characters using methods such as n-gram models, Byte Pair Encoding~\citep{gage1994new}, Word2Vec~\citep{mikolov2013efficient, mikolov2013distributed}, GloVe~\citep{pennington2014glove}, etc.
These methods allow language to be represented by a series of 1D vectors that serve as the input to the transformer.
Clearly a finite tokenization and such NLP embeddings are not directly applicable to physics, thus we  propose our own embedding method designed specifically for dynamical systems.
Consider learning the generalized mapping between the system's state space and embedded space: $\mathcal{F}: \mathbb{R}^{n\times d} \rightarrow \mathbb{R}^{e}$ and $\mathcal{G}: \mathbb{R}^{e} \rightarrow \mathbb{R}^{n\times d}$.
Naturally, multiple approaches can be used especially if the dimensionality of the embedded space is less than that of the state-space but this is not always the case.

The primary approach that we will propose is a Koopman observable embedding which is a technique that can be applied universally to all dynamical systems.
Considering the discrete time form of the dynamical system in  Eq.~(\ref{eq:pde}), the evolution of the state variables can be abstracted by $\bm{\phi}_{i+1} = \mathbb{F}\left(\bm{\phi}_{i}\right)$ for which $\mathbb{F}$ is the dynamic map from one time-step to the next.
The foundation of Koopman theory states any dynamical system can be represented in terms of an infinite dimensional linear operator acting on an infinite set of state observable functions, $g\left(\bm{\phi}_{i}\right)$, such that:
\begin{equation}
    \mathcal{K}g\left(\bm{\phi}_{i}\right) \triangleq g \circ \mathbb{F}\left(\bm{\phi}_{i}\right),
\end{equation}
where $\mathcal{K}$ is the infinite-dimensional linear operator referred to as the Koopman operator~\citep{koopman1931hamiltonian}.
This implies that the system of observables can be evolved in time through repeated application of the Koopman operator:
\begin{equation}
    g\left(\bm{\phi}_{i+1}\right) = \mathcal{K} g\left(\bm{\phi}_{i}\right), \quad g\left(\bm{\phi}_{i+2}\right) = \mathcal{K}^{2} g\left(\bm{\phi}_{i}\right), \quad g\left(\bm{\phi}_{i+3}\right) = \mathcal{K}^{3} g\left(\bm{\phi}_{i}\right), \ldots
\end{equation}
in which $\mathcal{K}^{n}$ denotes a $n$-fold composition, e.g. $\mathcal{K}^{3}(g) = \mathcal{K} (\mathcal{K} (\mathcal{K}(g)))$.
Modeling the dynamics of a system through the linear Koopman space can be attractive due to its simplification of the dynamics but also the potential physical insights it brings along with it.
Spectral analysis of the Koopman operator can reveal fundamental dynamical modes that drive the system's evolution in time.

Koopman  theory can be viewed as a trade off between lifting the state space into observable space with more complex states but simpler dynamics.
In practice, the Koopman operator must be finitely approximated.
This finite approximation requires the identification of the essential measurement functions that govern the system's dynamics and the respective approximate Koopman operator.
Data-driven machine learning has proven to be an effective approach for learning key Koopman observables for  modeling, control and dynamical mode analysis of many physical systems~\citep{li2017extended, korda2018linear, korda2020data, mezic2020numerical}.
In recent years, the use of deep neural networks for learning Koopman dynamics has proven to be successful~\cite{takeishi2017learning, lusch2018deep, otto2019linearly, brunton2021modern}.
While deep learning methods have enabled greater success with discovering Koopman observables and operators such approaches have yet to be demonstrated for the long time prediction of high-dimensional systems.
This is likely due to the approximation of the  finite-dimensional Koopman observables, limited data and complete dependence on the discovered Koopman operator $\mathcal{K}$ to model the dynamics.
Suggesting the prediction of a system through a single linear transform clearly has significant limitations and is fundamentally a naive approach from a machine learning perspective.
\begin{figure}[h]
    \centering
    \includegraphics[width=0.9\textwidth]{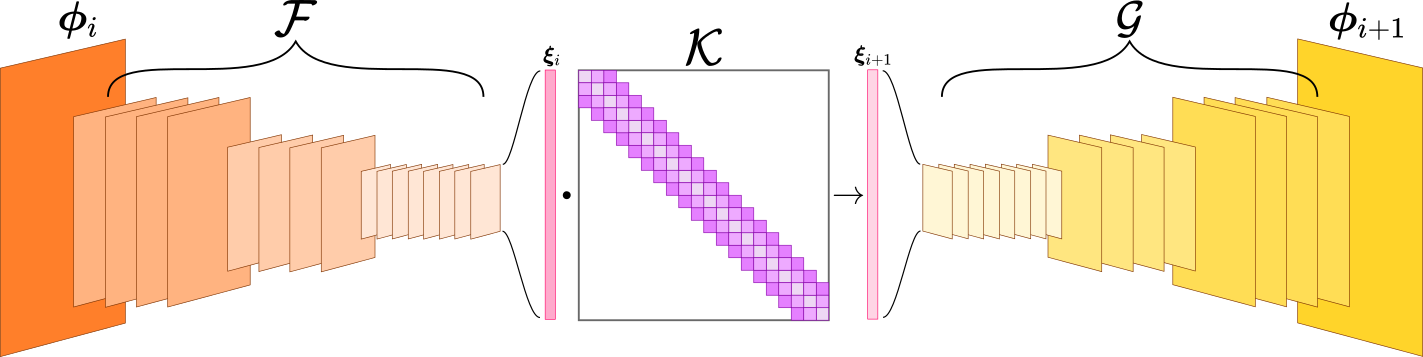}
    \caption{Example of a Koopman embedding for a two-dimensional system using a convolution encoder-decoder model.
    The encoder model, $\mathcal{F}$, projects the physical states into the approximate Koopman observable embedding.
    The decoder model, $\mathcal{G}$ recovers the physical states from the embedding.}
    \label{fig:koopman}
\end{figure}

In this work, we propose using approximate Koopman dynamics as a methodology to develop embeddings for the transformer model such that $\mathcal{F}(\bm{\phi}_{i}) \triangleq g(\bm{\phi}_{i})$.
As seen in Fig.~\ref{fig:koopman}, the embedding model follows a standard encoder-decoder model with the middle latent variables being the Koopman observables.
In this model, the Koopman operator assumes the form of a learnable banded matrix that is optimized with the auto-encoder.
Imposing some level of inductive bias on the form of the Koopman matrix is fairly common in similar deep Koopman works~\citep{lusch2018deep, otto2019linearly, i2020Learning}.
We found that this reduction of learnable parameters helped encourage the model to discover better dynamical modes, preventing the model from overfitting to high-frequency fluctuations.
Additionally, this form requires significantly less memory to store allowing models with embedding vectors of higher-dimensionality to be trained.
This learned Koopman operator is disposed of once training of the embedding model is complete.
Given the data set of physical state time-series, $\mathcal{D}_{\Phi}=\left\{\Phi^{i}\right\}_{i}^{D}$, the Koopman embedding model is trained with the following loss:
\begin{equation}
    \mathcal{L}_{\mathcal{D}_{\Phi}} = \sum_{i=1}^{D}\sum_{j=0}^{T} \lambda_{0} \underbrace{MSE\left(\bm{\phi}^{i}_{j},  \mathcal{G}\circ\mathcal{F}\left(\bm{\phi}^{i}_{j}\right)\right)}_{Reconstruction} + \lambda_{1} \underbrace{MSE\left(\bm{\phi}^{i}_{j}, \mathcal{G}\circ\mathcal{K}^{j}\mathcal{F}\left(\bm{\phi}^{i}_{0}\right)\right)}_{Dynamics} + \lambda_{2}\underbrace{\norm{\mathcal{K}}^{2}_{2}}_{Decay}.
\end{equation}
This loss function consists of three components: the first is a reconstruction loss which ensures a consistent mapping to and from the embedded representation. 
The second is the Koopman dynamics loss which pushes $\bm{\xi}_{j}$ to follow linear dynamics. 
The last term decays the Koopman operator's parameters to help force the model to discover meaningful dynamical modes and further prevent overfitting. 

In reference to traditional NLP embeddings, we believe our Koopman observable embedding has a motivation similar to Word2Vec~\citep{mikolov2013efficient} as well as more recent embedding methods such as context2vec~\citep{melamud2016context2vec}, ELMo~\citep{peters2018deep}, etc. 
These methods are based on word context and association to develop a map where words that are related or synonymous to each other have similar embedded vectors.
The Koopman embedding model has a similar objective encouraging physical realizations containing similar dynamical modes to also have similar embeddings.
This is because the Koopman operator is time-invariant which means the embedded states must share the same basis functions that govern their evolution.
As a result, the loss function of the embedding model rewards time-steps that are near each other in time or have the same underlying dynamics to have similar embeddings.
Hence, our goal with the embedding model is to not find the true Koopman observables or operator, but rather leverage Koopman to enforce physical context and association using the learned dynamical modes.

\section{Experiments and Results}
\noindent
The proposed transformer model is implemented for three dynamical systems.
The classical Lorenz system is used as a baseline test case in Section~\ref{sec:lorenz} to compare the proposed model to alternative machine learning approaches due to the Lorenz system's numerical sensitivity.
Following in Section~\ref{sec:cylinder}, we consider the modeling of two-dimensional Navier-Stokes fluid flow and compare different embedding methods for the transformer.
Lastly, to demonstrate the models scalability, we demonstrate in Section~\ref{sec:gray_scott} the transformer for the prediction of a three-dimensional reaction-diffusion system.
These examples encompass surrogate modeling physical systems with stochastic initial conditions and parameters.
\label{sec:numerical}
\subsection{Chaotic Dynamics}
\noindent
\label{sec:lorenz}
As a foundational numerical example to rigorously compare the proposed model to other classical machine learning techniques, we will first look at surrogate modeling of the Lorenz system governed by:
\begin{equation}
    \frac{dx}{dt} = \sigma \left(y - x \right), \quad \frac{dy}{dt} = x \left(\rho - z \right) - y, \quad \frac{dz}{dt} = xy - \beta z.
\end{equation}
We use the classical parameters of $\rho = 28, \sigma = 10, \beta = 8/3$.
For this numerical example, we wish to develop a surrogate model for predicting the Lorenz system given a random initial state  $x_{0}\sim\mathcal{U}(-20, 20)$, $y_{0}\sim\mathcal{U}(-20, 20)$ and $z_{0}\sim\mathcal{U}(10, 40)$.
In other words, we wish to surrogate model various initial value problems for this system of ODEs.
The Lorenz system is used because of its well known chaotic dynamics which make it extremely sensitive to numerical perturbations and thus an excellent benchmark for assessing a machine learning model's accuracy.

\begin{figure}[h]
    \centering
    \includegraphics[width=0.7\textwidth]{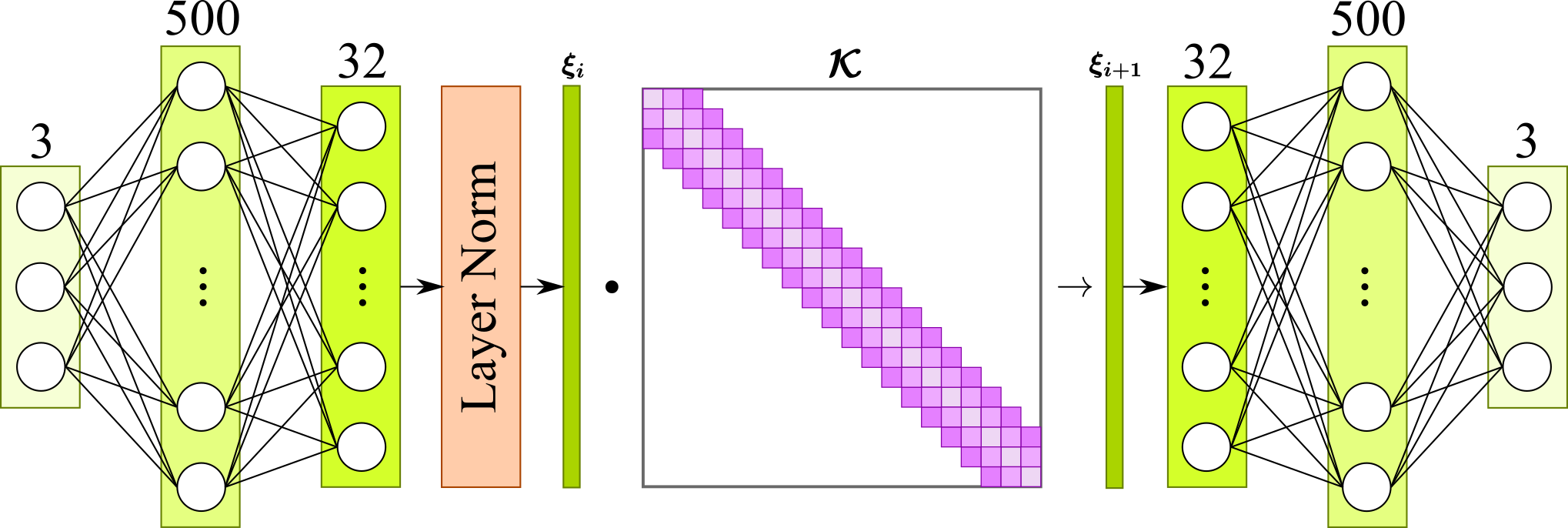}
    \caption{Fully-connected embedding network with ReLU activation functions for the Lorenz system.}
    \label{fig:koopman_model_lorenz}
\end{figure}

A total of four alternative machine learning models are implemented: a fully-connected auto-regressive model, a fully-connected LSTM model, a deep neural network Koopman model and lastly an echo-state model.
All of these types of models have been proposed in past literature for predicting various physical systems (e.g. auto-regressive~\cite{geneva2020modeling}, fully-connected LSTMs~\cite{zhao2019long}, deep Koopman~\citep{lusch2018deep, otto2019linearly} and echo-state models~\citep{chattopadhyay2019data, lukovsevivcius2012practical}).
Each are provided the same training, validation and testing data sets containing $2048$, $64$ and $256$ time-series at a time-step size of $\Delta t = 0.01$ solved using a Runge-Kutta numerical solver, respectively.
The training data set contains time-series of $256$ time-steps  while the validation and testing data sets have $1024$ time-steps.
Each model is allowed to train for $500$ epochs if applicable.
The proposed transformer and embedding model are trained for $200$ and $300$ epochs, respectively, with an embedding dimension of $32$.
The embedding model is a simple fully-connected encoder-decoder model, $\mathcal{F}: \mathbb{R}^{3} \rightarrow \mathbb{R}^{32}; \mathcal{G}: \mathbb{R}^{32} \rightarrow \mathbb{R}^{3}$, illustrated in Fig.~\ref{fig:koopman_model_lorenz}.
The transformer is trained with a context length of $64$ with $4$ transformer decoder layers.
Both the training and validation data sets were chunked into a set of $64$ steps for the alternative models, when applicable, to train on the same context length.

\begin{figure}[h]
    \centering
    \includegraphics[width=\textwidth]{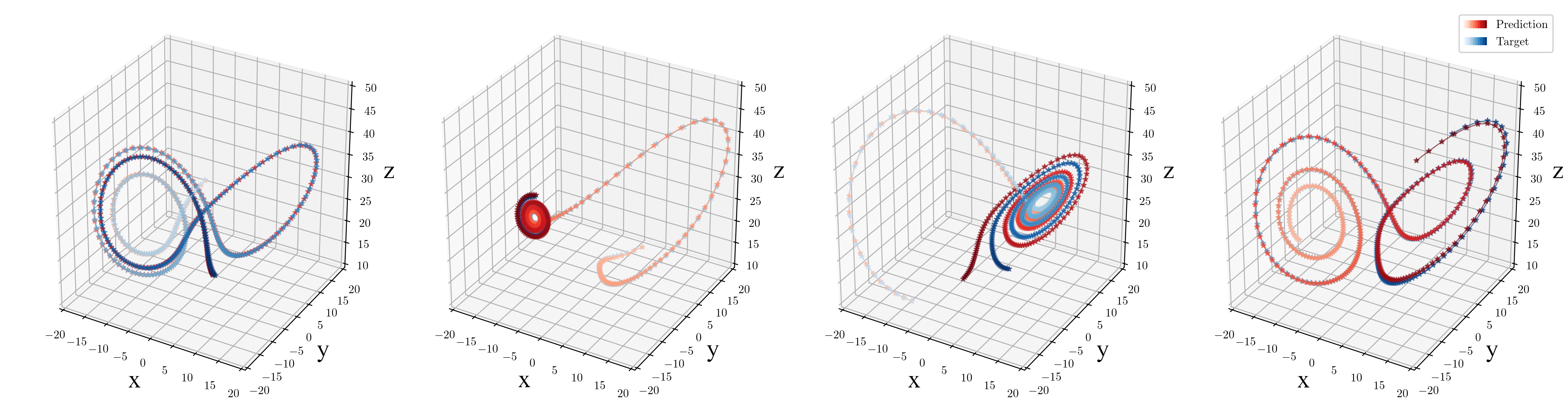}
    \caption{Four test case predictions using the transformer model for $320$ time-steps.}
    \label{fig:lorenz_pred}
\end{figure}

\begin{figure}[h]
    \centering
    \begin{subfigure}{0.4\textwidth}
        \centering
        \includegraphics[height=3.5cm]{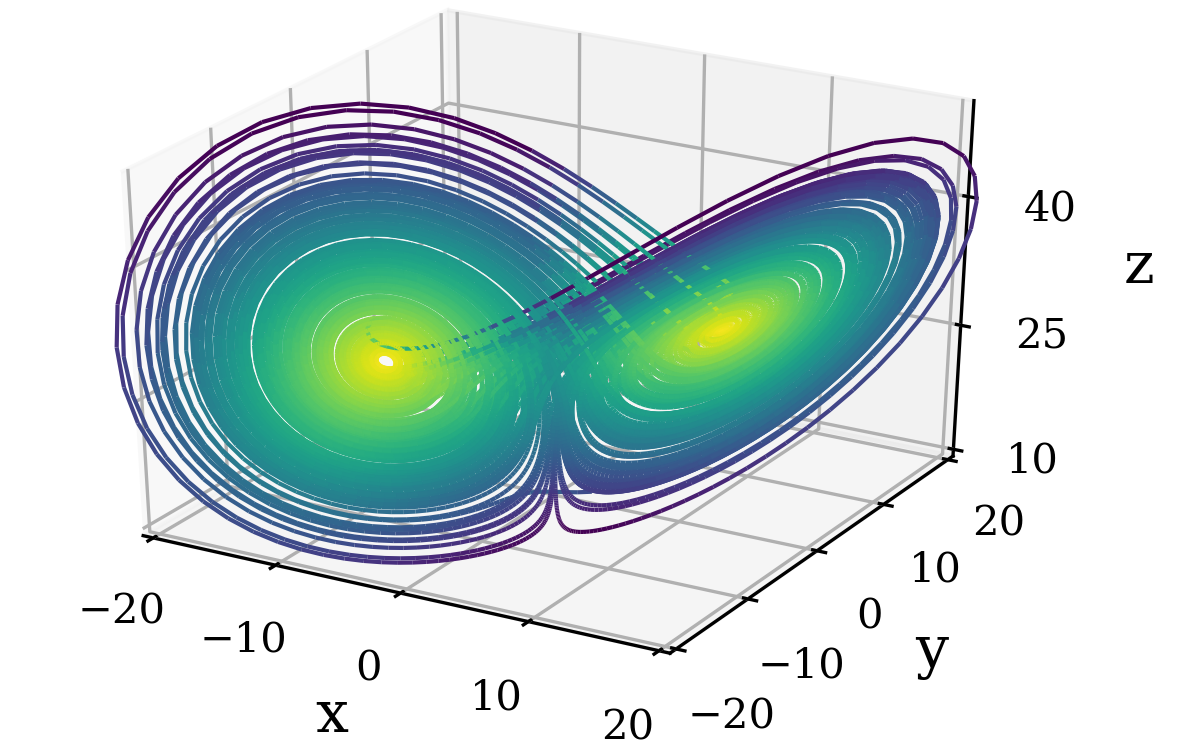}
        \caption{Numerical Solver}
    \end{subfigure}
    ~
    \begin{subfigure}{0.4\textwidth}
        \centering
        \includegraphics[height=3.5cm]{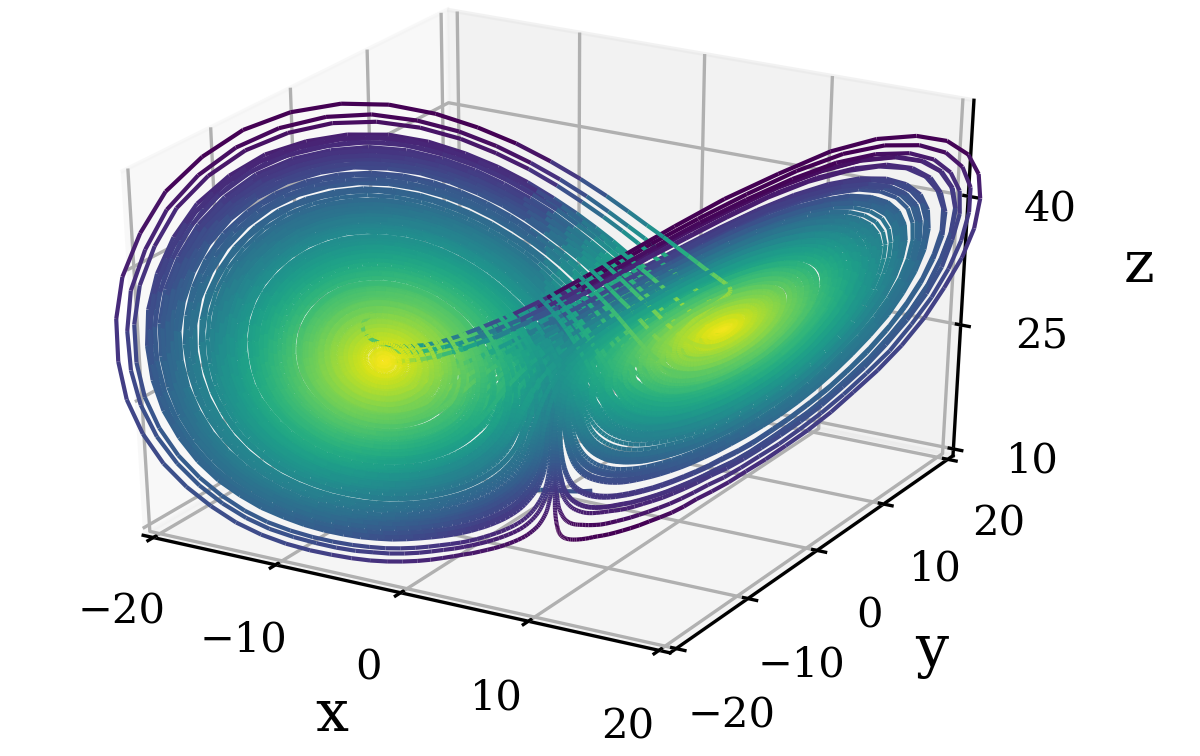}
        \caption{Transformer}
    \end{subfigure}
    \caption{Lorenz solution of $25$k time-steps with $\Delta t = 0.01$.}
    \label{fig:lorenz_manifold}
\end{figure}

We plot four separate test cases in Fig.~\ref{fig:lorenz_pred} for which only the initial state is provided and the transformer model predicts $320$ time-steps.
Several test predictions for alternative models are provided in~\ref{app:lorenz}.
In general, we can see that the transformer model is able to yield extremely accurate predictions even beyond its context length.
Additionally, we plot the Lorenz solution for $25$k time-steps from a numerical solver and predicted from the transformer model in Fig.~\ref{fig:lorenz_manifold}. 
Note that both have the same structure, which qualitatively indicates that the transformer indeed maintains physical dynamics.

\begin{figure}[h]
    \centering
    \begin{subfigure}{0.45\textwidth}
        \centering
        \includegraphics[height=5cm]{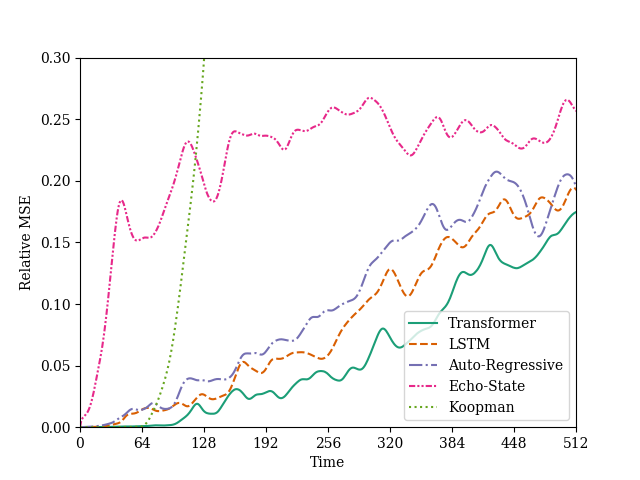}
        \caption{}
        \label{fig:lorenz_error}
    \end{subfigure}
     ~
    \begin{subfigure}{0.45\textwidth}
        \centering
        \includegraphics[height=5cm]{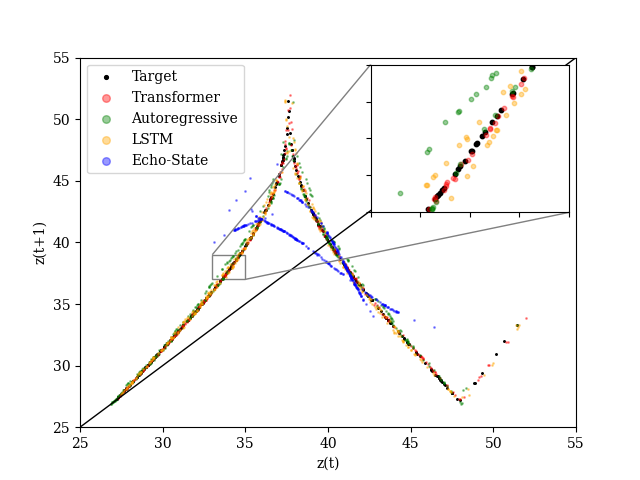}
        \caption{}
        \label{fig:lorenz_map}
    \end{subfigure}
    \caption{(a) The test relative mean-squared-error (MSE) with respect to time. (b) The Lorenz map produced by each model.}
\end{figure}

\begin{table}[h]
    \centering
    \caption{Test set relative mean-squared-error (MSE) for surrogate modeling the Lorenz system at several time-step intervals.}
    \begin{threeparttable}
    \begin{tabular}{  c | c | c | c | c }
       \multicolumn{1}{c}{} & & \multicolumn{3}{c}{Relative MSE}  \\ \hline
        Model & Parameters & $[0-64)$ & $[64-128)$ & $[128-192)$ \\ \hline \hline
        Transformer & $36$k/$54$k$^{\dagger}$ & $0.0003$ & $\bm{0.0060}$ & $\bm{0.0221}$ \\ \hline
        LSTM & $103$k & $0.0041$ & $0.0175$ & $0.0369$ \\ \hline
        Autoregressive & $92$k & $0.0057$ & $0.0253$ & $0.0485$ \\ \hline
        Echo State & $7.5$k/$6.3$m$^{\ddagger}$ & $0.1026$ & $0.1917$ & $0.2209$ \\ \hline
        Koopman & $108$k & $\bm{0.0001}$ & $0.0962$ & $2.0315$ \\
    \end{tabular}
    \begin{tablenotes}\footnotesize
    \item $^{\dagger}$ Learnable parameters for the embedding/transformer model.
    \item $^{\ddagger}$ Learnable output parameters/fixed input and reservoir parameters.
    \end{tablenotes}
    \end{threeparttable}
    \label{table:lorenz_error}
\end{table}

The proposed transformer and alternative models' relative mean squared errors for the test set are plotted in Fig.~\ref{fig:lorenz_error} as a function of time and listed in Table~\ref{table:lorenz_error} segmented into several intervals based on the transformer's context length.
In general, we can see all deep learning models perform well in the time-series length for which they were trained with the deep Koopman model performing the best followed by the transformer.
As we extrapolate our predictions past the trained context range, the benefits of the transformer become apparent with it achieving the best accuracy for later times.
To quantify accuracy of the chaotic dynamics of each model, the Lorenz map is plotted in Fig.~\ref{fig:lorenz_map} which is a well-defined relation between successive $z$ local maxima despite the Lorenz's chaotic nature.
Calculated using $25$k time-step predictions from each model, again we can see that the transformer model agrees the best with the numerical solver indicating that it has learned the best physical dynamics of all the tested models.

Additionally, we test the transformer's sensitivity to contaminated data by adding white noise to the training observations scaled by the magnitude of the state variables.
Each model tested with clean data is retrained with data perturbed by $1$\% and $5$\% noise.
The effects of these two different noise levels is qualitatively illustrated in~\ref{app:lorenz}.
The errors are listed in Table~\ref{table:lorenz_error_noise}.
In general, we can indeed see that the transformer can still perform adequately with noisy data by being the best performing model for $1$\% noise and still being competitive, particularly at later time-steps, with $5$\% noise.

\begin{table}[h]
    \centering
    \caption{Test set relative mean-squared-error (MSE) for surrogate modeling the Lorenz system at several time-step intervals with noisy data.}
    \begin{tabular}{  c | c | c | c || c | c | c }
        & \multicolumn{3}{c||}{Relative MSE 1\% Noise} & \multicolumn{3}{c}{Relative MSE 5\% Noise}\\ \hline
        Model & $[0-64)$ & $[64-128)$ & $[128-192)$ & $[0-64)$ & $[64-128)$ & $[128-192)$ \\ \hline \hline
        Transformer & $\bm{0.0021}$ & $\bm{0.0216}$ & $\bm{0.0429}$ & $0.0210$ & $0.0759$ & $\bm{0.1292}$\\ \hline
        LSTM & $0.0045$ & $0.0218$ & $0.0437$ & $0.0212$ & $\bm{0.0758}$ & $0.1324$\\ \hline
        Autoregressive & $0.0114$ & $0.0417$ & $0.0901$ & $0.0760$ & $0.2060$ & $0.2065$\\ \hline
        Echo State & $0.0859$ & $0.1686$ & $0.2102$ & $0.1000$ & $0.1581$ & $0.2051$\\ \hline
        Koopman & $0.0047$ & $0.1192$ & $0.1597$ & $\bm{0.0200}$ & $0.0787$ & $0.1906$\\
    \end{tabular}
    \label{table:lorenz_error_noise}
\end{table}

\begin{figure}[h]
    \centering
    \includegraphics[width=0.95\textwidth,trim={3cm 0 1cm 0},clip]{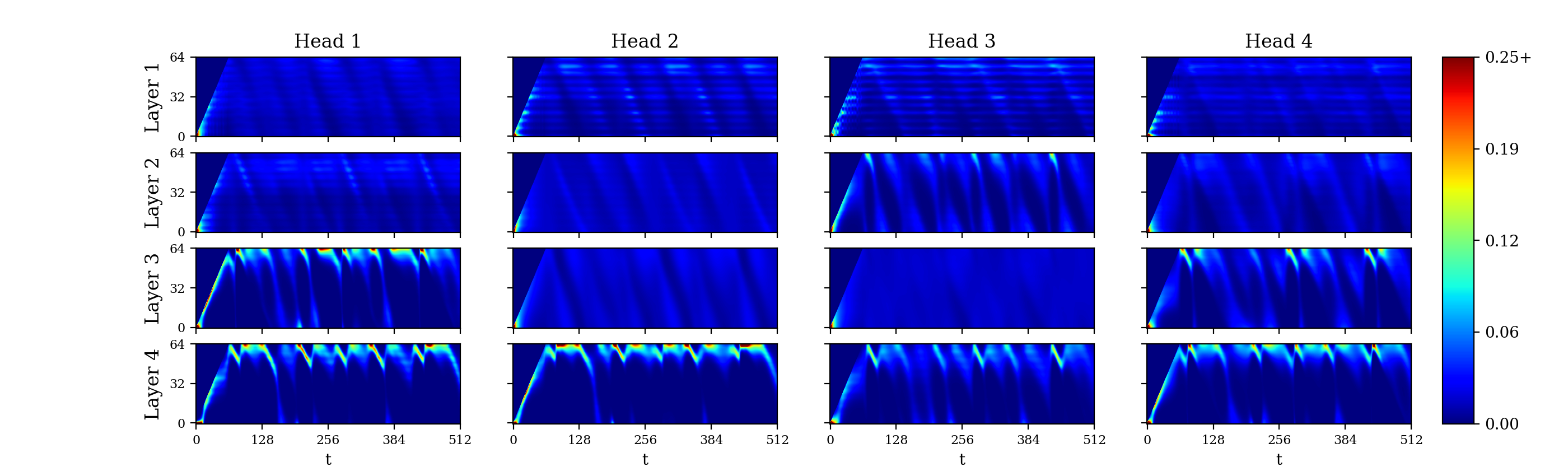}
    \includegraphics[width=\textwidth]{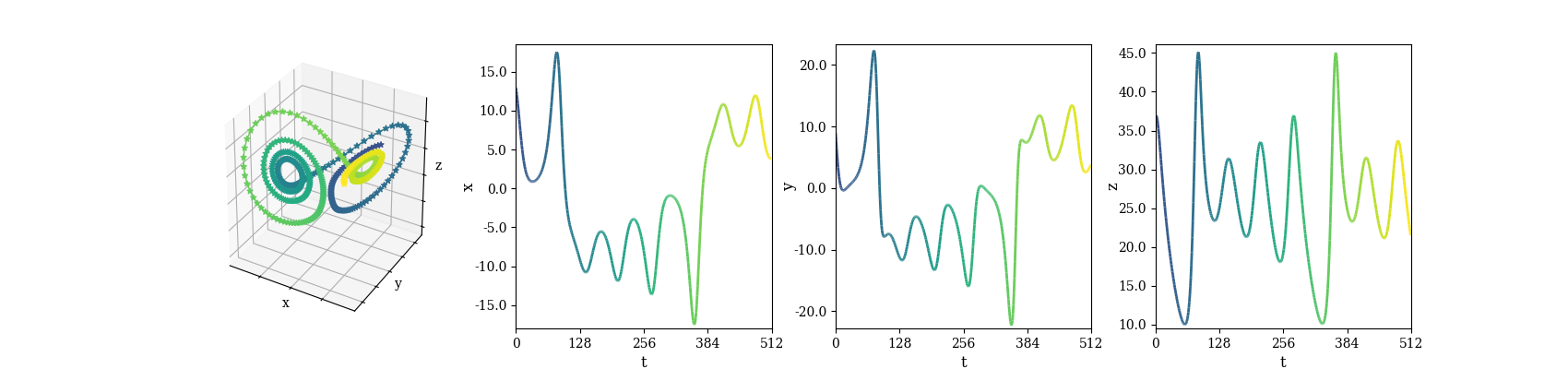}
    \caption{Attention vectors of the transformer model for the prediction of a single test case of the Lorenz system. The top contours illustrate the attention weights at each time-step; the bottom shows the respective state variables.}
    \label{fig:lorenz_attention}
\end{figure}

The self-attention vectors, $\bm{\alpha}_{i}\in \mathbb{R}^{64}$, for a single time-series prediction of $512$ steps are plotted in Fig.~\ref{fig:lorenz_attention}.
For time-steps over $i \ge 64$, the full context length of the transformer is used with the attention weight $\alpha_{i,64}$ corresponding to the most recent time-step. 
Note that the transformer has learned multi-scale temporal dependencies not achievable with other machine learning architectures.
Additionally, this verifies that each attention head is learning different dependencies suggesting that each head may be a particular component of the transformer's latent dynamics.
This aligns with what is observed in NLP when using multi-head attention which increases the transformer's predictive accuracy~\cite{vaswani2017attention}.

\subsection{2D Fluid Dynamics}
\label{sec:cylinder}
\noindent
The next dynamical system we will test is transient 2D fluid flow governed by the Navier-Stokes equations:
\begin{equation}
    \frac{\partial u_{i}}{\partial t} + u_{j}\frac{\partial u_{i}}{\partial x_{j}} = -\frac{1}{\rho}\frac{\partial p}{\partial x_{i}} + \nu\frac{\partial^{2} u_{i}}{\partial x_{j}\partial x_{j}},
\end{equation}
in which $u_{i}$ and $p$ are the velocity and pressure, respectively.
$\nu$ is the viscosity of the fluid.
We consider modeling the classical problem of flow around a cylinder at various Reynolds number defined by $Re = u_{in}d/\nu$ in which $u_{in}=1$ and $d=2$ are the inlet velocity and cylinder diameter, respectively.
In this work, we choose to develop a surrogate model to predict the solution at any Reynolds number between $Re \sim\mathcal{U}\left(100, 750\right)$.
This problem is a fairly classical flow to investigate in scientific machine learning with various levels of difficulty~\citep{lee2017prediction, morton2018deep, lusch2018deep, han2019novel, xu2020multi, geneva2020multi}.
Here we choose one of the more difficult forms: model the flow starting at a steady state flow field at $t=0$.
Meaning the model is provided zero information on the structure of the cylinder wake during testing other than the viscosity.

Training, validation and testing data is obtained using the OpenFOAM simulator~\citep{jasak2007openfoam}, from which a rectangular structured sub-domain is sampled centered around the cylinder's wake.
Given that the flow is two-dimensional, our model will predict the $x$-velocity, $y$-velocity and pressure fields, $\left(\bm{u}_{x}, \bm{u}_{y}, \bm{p}\right)\in\mathbb{R}^{3 \times 64 \times 128}$.
The training, validation and test data sets consist of $27$, $6$ and $7$ fluid flow simulations, respectively with $400$ time-steps each at a physical time-step size of $\Delta t = 0.5$.
As a base line model, we train a convolutional encoder-decoder model with a stack of three convolutional LSTMs~\citep{xingjian2015convolutional} in the center.
The input for this model consists of the velocity fields, pressure field and viscosity of the fluid.
We note that convolutional LSTMs have been used extensively in recent scientific machine learning literature for modeling various physical systems including fluid dynamics~\citep{han2019novel, wiewel2019latent, tang2020deep, geneva2020multi, maulik2021reduced}, thus can be considered a state-of-the-art approach.
The convolutional LSTM model is trained for $500$ epochs.

\begin{figure}[h]
    \centering
    \includegraphics[width=\textwidth]{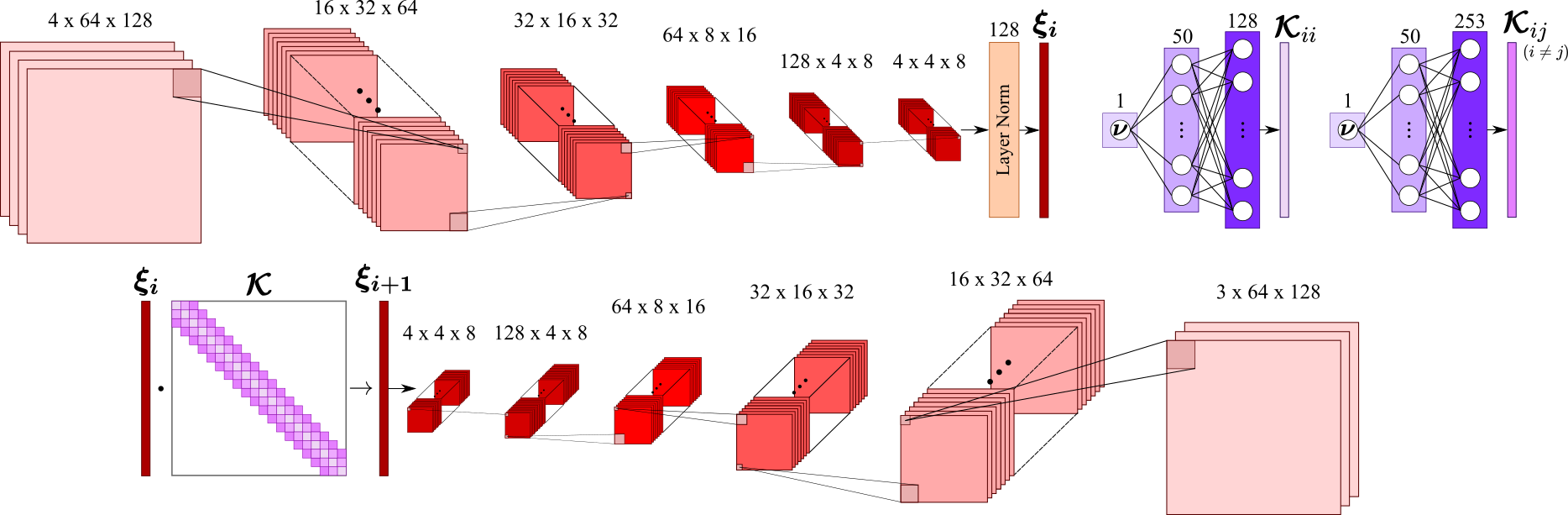}
    \caption{2D convolutional embedding network with ReLU activation functions for the flow around a cylinder system consisting of $5$ convolutional encoding/decoding layers. Each convolutional operator has a kernel size of $(3,3)$. In the decoder, the feature maps are up-sampled before applying a standard convolution. Additionally, two auxiliary fully-connected networks are used to predict the diagonal and off-diagonal elements of the Koopman operator for each viscosity $\bm{\nu}$.}
    \label{fig:koopman_model_cylinder}
\end{figure}

\begin{figure}[h]
    \centering
    \begin{subfigure}{\textwidth}
        \centering
        \includegraphics[width=\textwidth]{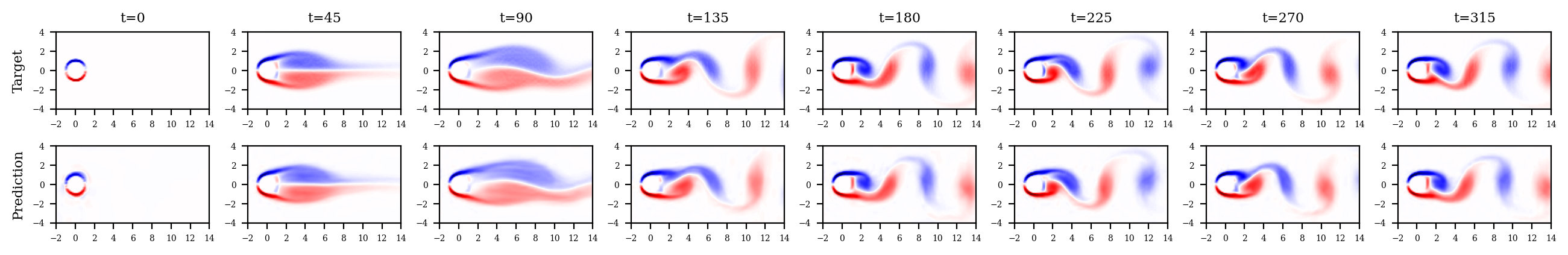}
    \end{subfigure}\\
    \begin{subfigure}{\textwidth}
        \centering
        \includegraphics[width=\textwidth]{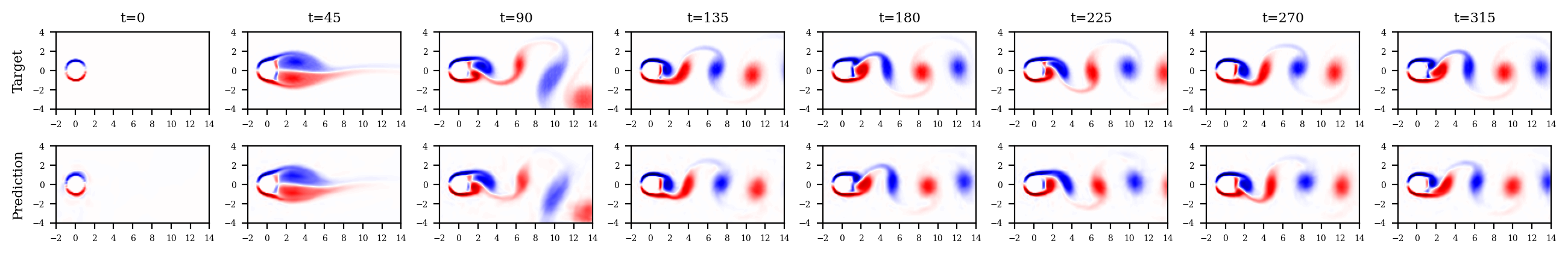}
    \end{subfigure}
    \caption{Vorticity, $\bm{\omega} = \nabla_{x}\bm{u}_{y} - \nabla_{y}\bm{u}_{x}$, of two test case predictions using the proposed transformer with Koopman embeddings at Reynolds numbers $233$ (top) and $633$ (bottom).}
    \label{fig:cylinder_vort}
\end{figure}

\begin{figure}[h]
    \centering
    \includegraphics[width=\textwidth]{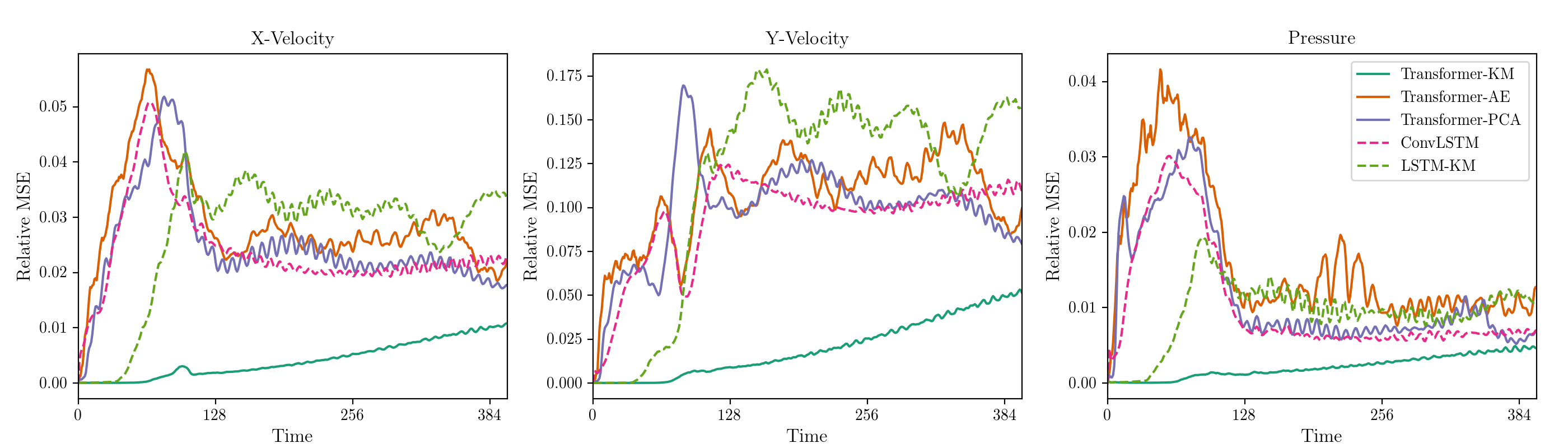}
    \caption{Test set relative mean-squared-error (MSE) of the transformer with Koopman (KM), auto-encoder (AE) and PCA embedding methods, the convolutional LSTM model (ConvLSTM) and fully-connected LSTM with Koopman embeddings (LSTM-KM).}
    \label{fig:cylinder_error}
\end{figure}

Additionally, three different embedding methods are implemented: the first is the proposed Koopman observable embedding using a convolutional auto-encoder illustrated in Fig.~\ref{fig:koopman_model_cylinder}. This model encodes the fluid $x$-velocity, $y$-velocity, pressure and viscosity fields to an embedded dimension of $128$,  $\mathcal{F}: \mathbb{R}^{4\times64\times128} \rightarrow \mathbb{R}^{128}; \mathcal{G}: \mathbb{R}^{128} \rightarrow \mathbb{R}^{3 \times64\times128}$.
The second embedding method is using the same convolutional auto-encoder model but without the enforcement of Koopman dynamics on the embedded variables.
The third embedding method tested was principal component analysis (PCA) as a classical baseline.
For each embedding method an identical transformer model with a context length of $128$ and $4$ transformer decoder layers is trained.
Similar to the previous example, the embedding models are trained for $300$ epochs when applicable and the transformer is trained for $200$.
To further isolate the impact of the transformer model from the embedding model, we also train a fully-connected model with $4$ LSTM cells using the Koopman embedding as an input for $200$ epochs.
Each trained model is tested on the test set by providing the initial laminar state at $t=0$ with the fluid viscosity and allowing the model to predict $400$ time-steps into the future.
Two test predictions using the proposed transformer model with Koopman embeddings are plotted in Fig.~\ref{fig:cylinder_vort} in which the predicted vorticity fields are in good agreement with the true solution.
The test set relative mean square error for each output field for each model is plotted in Fig.~\ref{fig:cylinder_error}.
The errors of each field over the entire time-series are listed in Table~\ref{table:cylinder_error}.
Additional results are provided in~\ref{app:cylinder}.

\begin{table}[ht]
    \centering
    \caption{Test set relative mean-squared-error (MSE) of each output field for surrogate modeling 2D fluid flow past a cylinder. Models listed include the transformer with Koopman (KM), auto-encoder (AE) and PCA embedding methods, the convolutional LSTM model (ConvLSTM) and fully-connected LSTM with Koopman embeddings (LSTM-KM).}
    \begin{threeparttable}
    \begin{tabular}{  c | c | c | c | c }
       \multicolumn{1}{c}{} & & \multicolumn{3}{c}{Relative MSE $[0-400]$} \\ \hline
        Model & Parameters & $\bm{u}_{x}$ & $\bm{u}_{y}$ & $\bm{p}$ \\ \hline \hline
        Transformer-KM & $224$k/$628$k$^{\dagger}$ & $\bm{0.0042}$ & $\bm{0.0198}$ & $\bm{0.0021}$ \\ \hline
        Transformer-AE & $224$k/$628$k$^{\dagger}$ & $0.0288$ & $0.1068$ & $0.0161$ \\ \hline
        Transformer-PCA & $3.1$m/$628$k$^{\dagger}$ & $0.0247$ & $0.0984$ & $0.0117$ \\ \hline
        ConvLSTM & $934$k & $0.0240$ & $0.0938$ & $0.0103$ \\ \hline
        LSTM-KM & $224$k/$759$k$^{\ddagger}$ & $0.0266$ & $0.1155$ & $0.0094$ \\
    \end{tabular}
    \begin{tablenotes}\footnotesize
    \item $^{\dagger}$ Learnable parameters for the embedding/ transformer model.
    \item $^{\ddagger}$ Learnable parameters for the embedding/ LSTM model.
    \end{tablenotes}
    \end{threeparttable}
    \label{table:cylinder_error}
\end{table}

\begin{figure}[h]
    \centering
    \includegraphics[width=\textwidth,trim={2cm 0 0cm 0},clip]{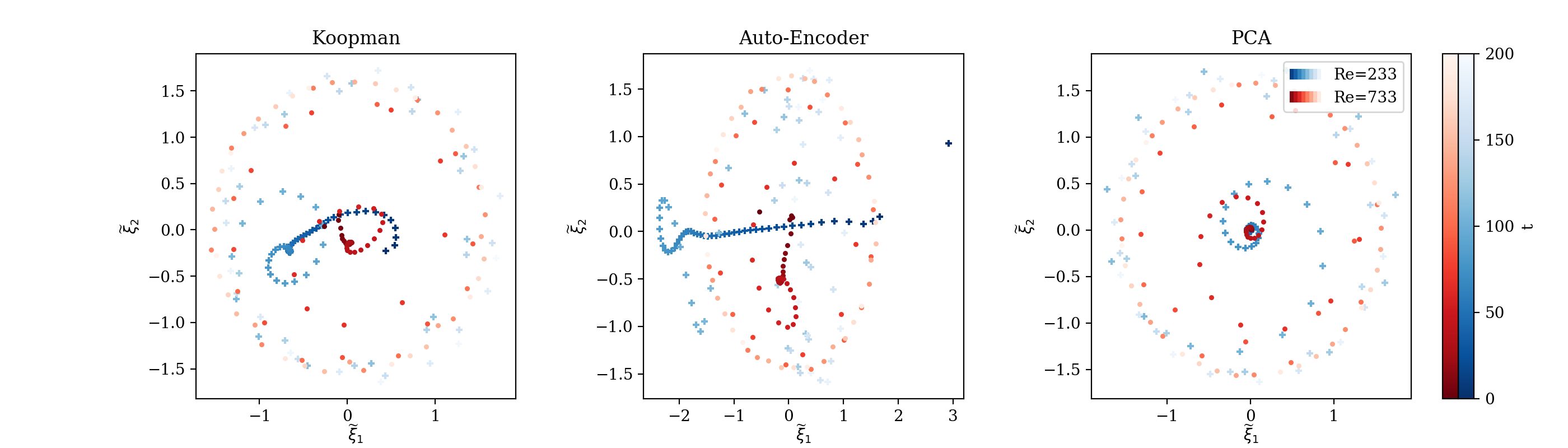}
    \caption{The two-dimensional principle subspace of the embedded vectors, $\bm{\xi}_{i}$, from each tested embedding model for two different Reynolds numbers.}
    \label{fig:cylinder_embedding}
\end{figure}

For all alternative models, a rapid error increase can be seen between $t=[0, 100]$ which is due to the transition from the laminar flow into vortex shedding.
This error then plateaus since each model is able to produce stable vortex shedding, as illustrated in Figs.~\ref{fig:cylinder_mag} \&~\ref{fig:cylinder_pressure} in~\ref{app:cylinder}.
The proposed transformer with Koopman embeddings is the only model that can accurately match the instantaneous states from the numerical solution.
These results indicate that the performance of the transformer is highly dependent on the embedding method used, which should be expected.
However, the transformer with self-attention is equally important to the model's success as indicated by the performance of the Koopman embedding LSTM model.
Compared to the widely used ConvLSTM model, the proposed transformer offers more reliable predictions for this fluid flow with less learnable parameters.
To gain a greater understanding of why the Koopman embedding performs better than the alternatives, we perform linear dimensionality reduction of the embedded states using PCA into two principle components, $\widetilde{\bm{\xi}}_{1}, \widetilde{\bm{\xi}}_{2}$, in Fig.~\ref{fig:cylinder_embedding}.
For all embedding methods, a circular structure is present reflecting the periodic dynamics of the vortex shedding.
Compared to the auto-encoder, we note that the Koopman principle subspace has a more consistent structure for later time-steps indicating that the Koopman loss term does indeed encourage the discovery of common dynamical modes.
Yet for early time-steps, the Koopman model has a unique trajectory between Reynolds numbers.
This lack of uniqueness with the PCA embedding at early time-steps results in the transformer not being able to differentiate between Reynolds numbers, lowering predictive accuracy with this embedding method.
The Koopman approach strikes a balance between structure but uniqueness between flows for the transformer to learn with.

\subsection{3D Reaction-Diffusion Dynamics}
\noindent
\label{sec:gray_scott}
The final numerical example to demonstrate the proposed transformer model is a 3D reaction-diffusion system governed by the Gray-Scott model:
\begin{equation}
    \frac{\partial u}{\partial t} = r_{u}\frac{\partial^{2} u}{\partial x_{i}^{2}} - uv^{2} + f\left(1-u\right), \quad \frac{\partial v}{\partial t} = r_{v}\frac{\partial^{2} v}{\partial x_{i}^{2}} + vu^{2} - \left(f + k\right)v,
\end{equation}
in which $u$ and $v$ are the concentration of two species, $r_{u}$ and $r_{v}$ are their respective diffusion rates, $k$ is the kill rate and $f$ is the feed rate.
This is a classical system of particular application to chemical processes as it models the following reaction: $U + 2V \rightarrow 3V$; $V\rightarrow P$.
For a set region of feed and kill rates, this seemingly simple system can yield a wide range of complex dynamics~\citep{pearson1993complex, lee1993pattern}.
Hence, under the right settings, this system is an excellent case study to push the proposed methodology to its predictive limits.
In this work, we will use the parameters: $r_{u}=0.2$, $r_{v}=0.1$, $k=0.055$ and $f=0.025$ which results in a complex dynamical reaction.
Akin to the first numerical example, the initial condition of this system is stochastic such that the system is seeded with $3$ randomly placed perturbations within the periodic domain.

\begin{figure}[h]
    \centering
    \includegraphics[width=\textwidth]{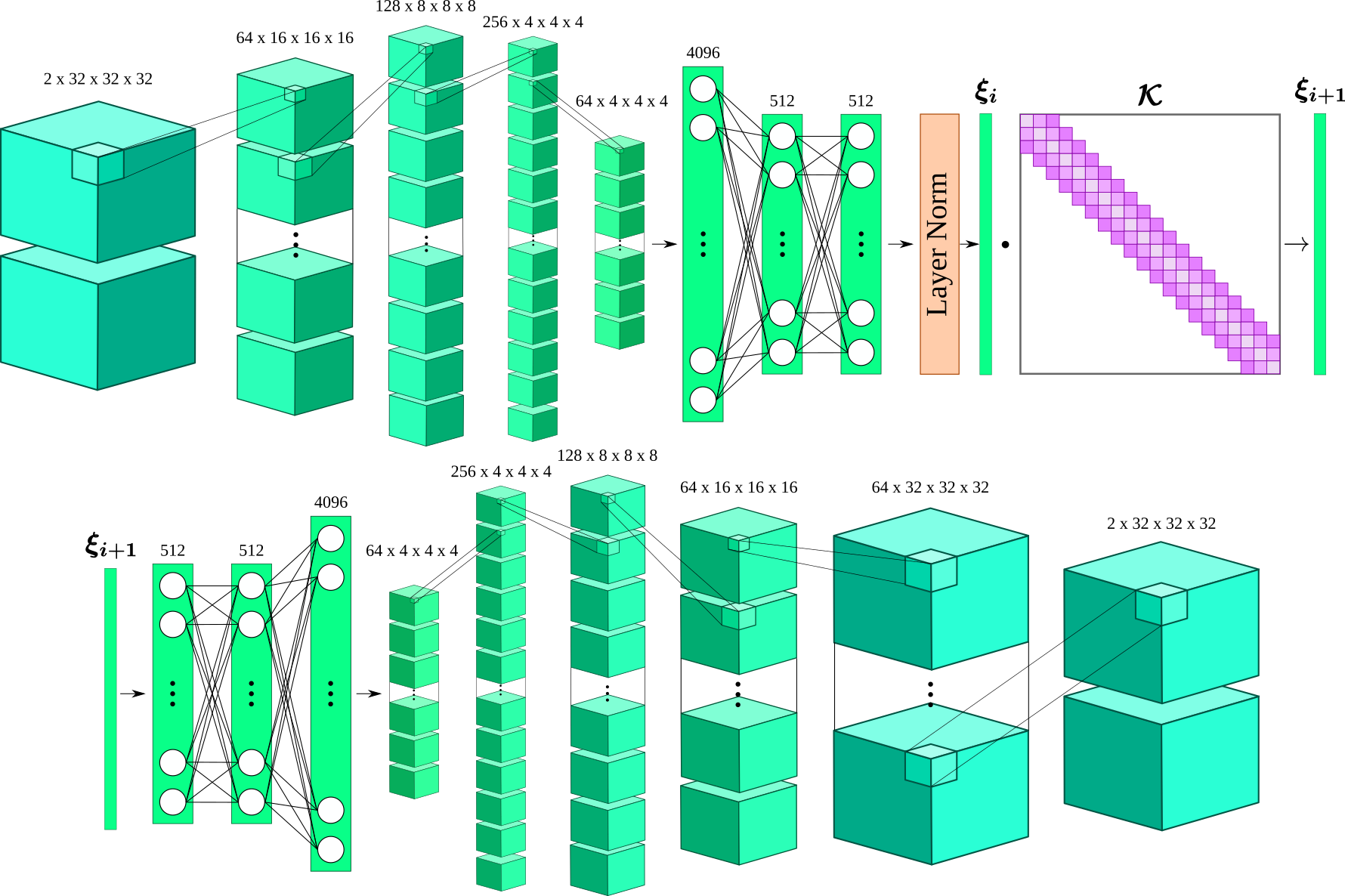}
    \caption{3D convolutional embedding network with leaky ReLU activation functions for the Gray-Scott system. Batch-normalization used between each of the convolutional layers. In the decoder, the feature maps are up-sampled before applying a standard 3D convolution.}
    \label{fig:koopman_model_gray}
\end{figure}

Training, validation and testing data are obtained from a Runge-Kutta finite difference simulation on a structured grid, $\left(\bm{u}, \bm{v}\right) \in \mathbb{R}^{2\times32\times32\times32}$.
The training, validation and test data sets consist of $512$, $16$ and $56$ time-series, respectively, with $200$ time-steps each at a physical time-step size of $\Delta t = 5$.
A 3D convolutional encoder-decoder is used to embed the two species into a $512$ embedding dimension,  $\mathcal{F}: \mathbb{R}^{2\times32\times32\times32} \rightarrow \mathbb{R}^{512}; \mathcal{G}: \mathbb{R}^{512} \rightarrow \mathbb{R}^{2\times32\times32\times32}$, illustrated in Fig.~\ref{fig:koopman_model_gray}.

\begin{figure}[h]
    \centering
    \centering
    \begin{subfigure}{\textwidth}
        \centering
        \includegraphics[width=0.95\textwidth]{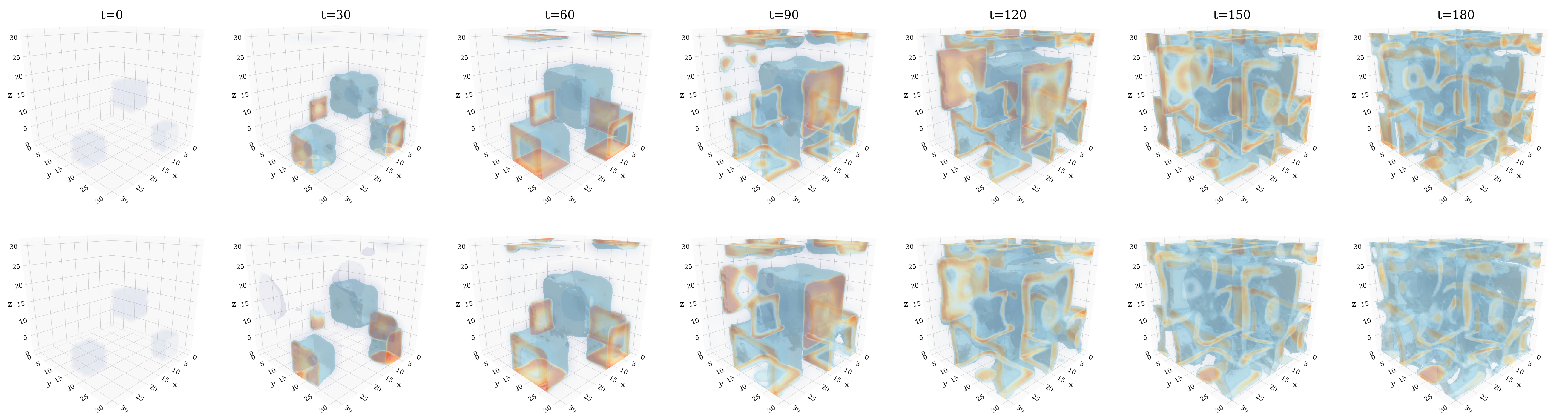}
        \caption{$\bm{u}$ target (top) and transformer prediction (bottom)}
    \end{subfigure}\\
    \begin{subfigure}{\textwidth}
        \centering
        \includegraphics[width=0.95\textwidth]{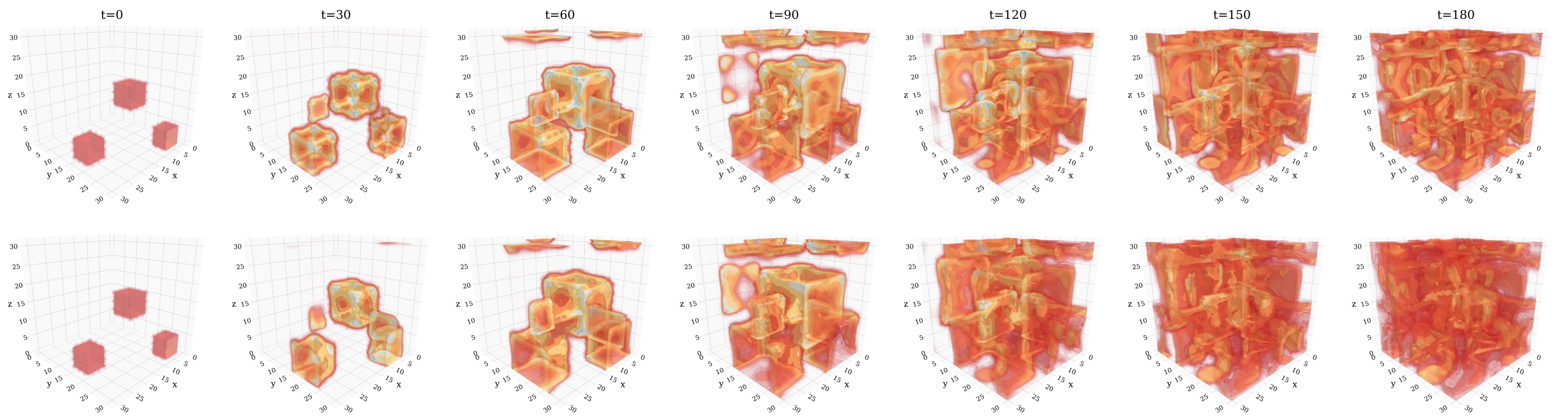}
        \caption{$\bm{v}$ target (top) and transformer prediction (bottom)}
    \end{subfigure}
    \caption{Test case volume plots for the Gray-Scott system. Isosurfaces displayed span the range $\bm{u},\bm{v}=[0.3,0.5]$ to show the inner structure.}
    \label{fig:gray_scott}
\end{figure}

Transformer models with varying depth are trained all with a context length of $128$.
All other model and training parameters for the transformer models are consistent.
A test prediction using the transformer model is shown in Fig.~\ref{fig:gray_scott} and the errors for each trained transformer are listed in Table~\ref{table:grayscott_error}.
Despite this system having complex dynamics in 3D space, the transformer is able to produce acceptable predictions with very similar structures as the numerical solver.
To increase the model's predictive accuracy we believe the limitation here is not in the transformer, but rather the number of training data and the inaccuracies of the embedding model due to the dimensionality reduction needed.
This is supported by the fact that increasing the transformer's depth does not yield considerable improvements for the test errors in Table~\ref{table:grayscott_error}.
Additional results are provided in~\ref{app:gray_scott}.
\begin{table}[h]
    \centering
    \caption{Test set relative mean-squared-error (MSE) for surrogate modeling 3D Gray-Scott system.}
    \begin{threeparttable}
    \newcolumntype{Y}{>{\centering\arraybackslash}X}
    \begin{tabularx}{0.75\textwidth}{c|c|c*{2}{|Y}}
       \multicolumn{3}{c|}{} & \multicolumn{2}{c}{Relative MSE $[0-200]$} \\ \hline
        Model & Layers & Parameters & $\bm{u}$ & $\bm{v}$\\ \hline \hline
        Transformer & $2$ & $6.2$m/$6.6$m$^{\dagger}$ & $0.0159$ & $0.0120$ \\ \hline
        Transformer & $4$ & $6.2$m/$12.9$m$^{\dagger}$ & $ 0.0154$ & $0.0130$ \\ \hline
        Transformer & $8$ & $6.2$m/$25.5$m$^{\dagger}$ & $\bm{0.0125}$ & $\bm{0.0101}$
    \end{tabularx}
    \begin{tablenotes}\footnotesize
    \item $^{\dagger}$ Learnable parameters for the embedding/ transformer model.
    \end{tablenotes}
    \end{threeparttable}
    \label{table:grayscott_error}
\end{table}
\section{Conclusion}
\label{sec:conclusion}
\noindent
While transformers and self-attention models have been established as a powerful framework for NLP tasks, the adoption of such methodologies has yet to fully permute other fields.
In this work, we have demonstrated the potential transformers have for modeling dynamics of physical phenomena.
The transformer architecture allows the model to learn longer and more complex temporal dependencies than alternative machine learning methods.
Such models can be particularly beneficial for physical systems that exhibit dynamics that evolve on multiple time scales or consist of multiple phases such as turbulent fluid flow, chemical reactions or molecular dynamics.
This can be attributed to the transformer's ability to draw information from many past time-steps directly with self-attention, learning accurate time integration without computationally expensive recurrent connections.

The key challenge of using the transformer model is identifying appropriate embeddings to represent the physical state of the system, for which we propose the use of Koopman dynamics to enforce dynamical context.
Using the proposed methods, our transformer surrogate can outperform alternative models widely seen in recent scientific machine learning literature.
The investigation of unsupervised pre-training of such transformer models as well as gaining a better understanding of what attention mechanisms imply for physical systems will be the subject  of works in the future.
\subsubsection*{Acknowledgments}
\noindent
The anonymous reviewers are thanked for their significant effort in improving and clarifying this manuscript. The work reported here was initiated  from the Defense Advanced Research Projects Agency (DARPA) under the Physics of Artificial Intelligence (PAI) program (contract HR$00111890034$).
The authors acknowledge computing resources provided by the AFOSR Office
of Scientific Research through the DURIP program and by the University of Notre Dame’s Center for Research Computing (CRC). 
The work of NG was supported by the National Science Foundation (NSF) Graduate Research Fellowship Program grant No. DGE-1313583.

\newpage
\bibliography{mybibfile}

\newpage
\appendix

\newpage
\section{Lorenz Supplementary Results}
\begin{figure}[h]
    \centering
    \includegraphics[width=0.98\textwidth]{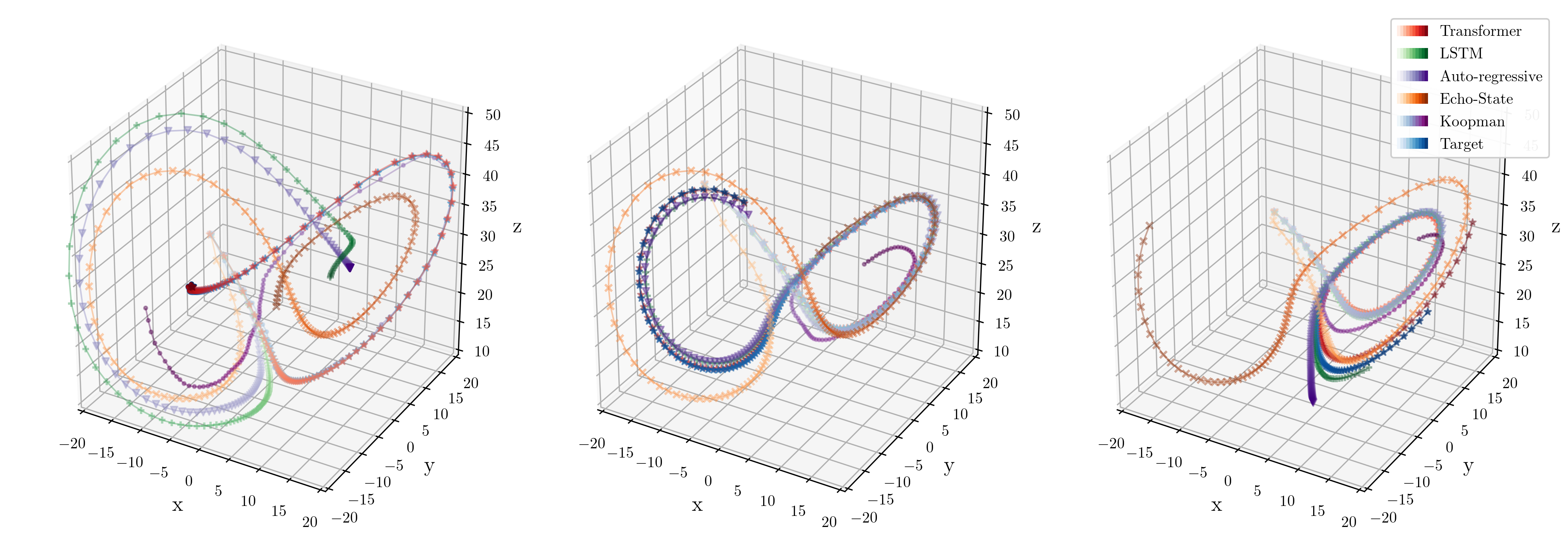}
    \caption{Three Lorenz test case predictions using each tested model for $128$ time-steps.}
    \label{fig:lorenz_models_pred}
\end{figure}

\begin{figure}[h]
    \centering
    \includegraphics[width=\textwidth]{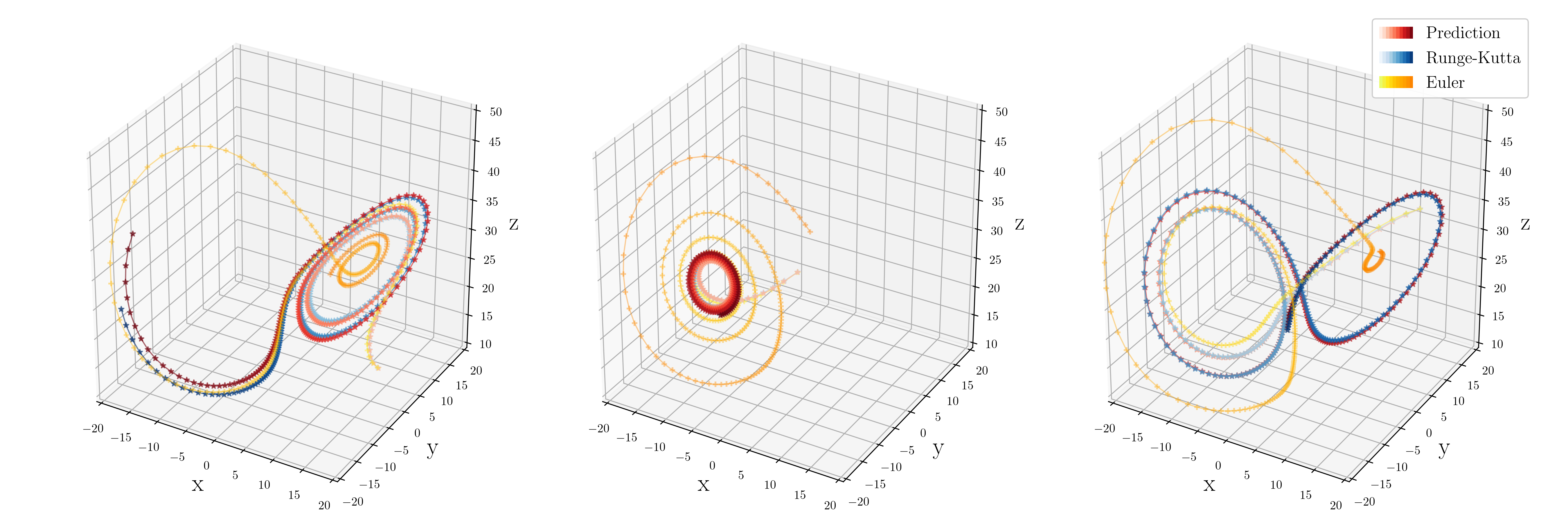}
    \caption{Three Lorenz test solutions solved using Runge-Kutta and Euler time-integration methods with the prediction of the transformer for $256$ time-steps.}
    \label{fig:lorenz_numeric_pred}
\end{figure}

\begin{figure}[h]
    \centering
    \includegraphics[width=0.8\textwidth]{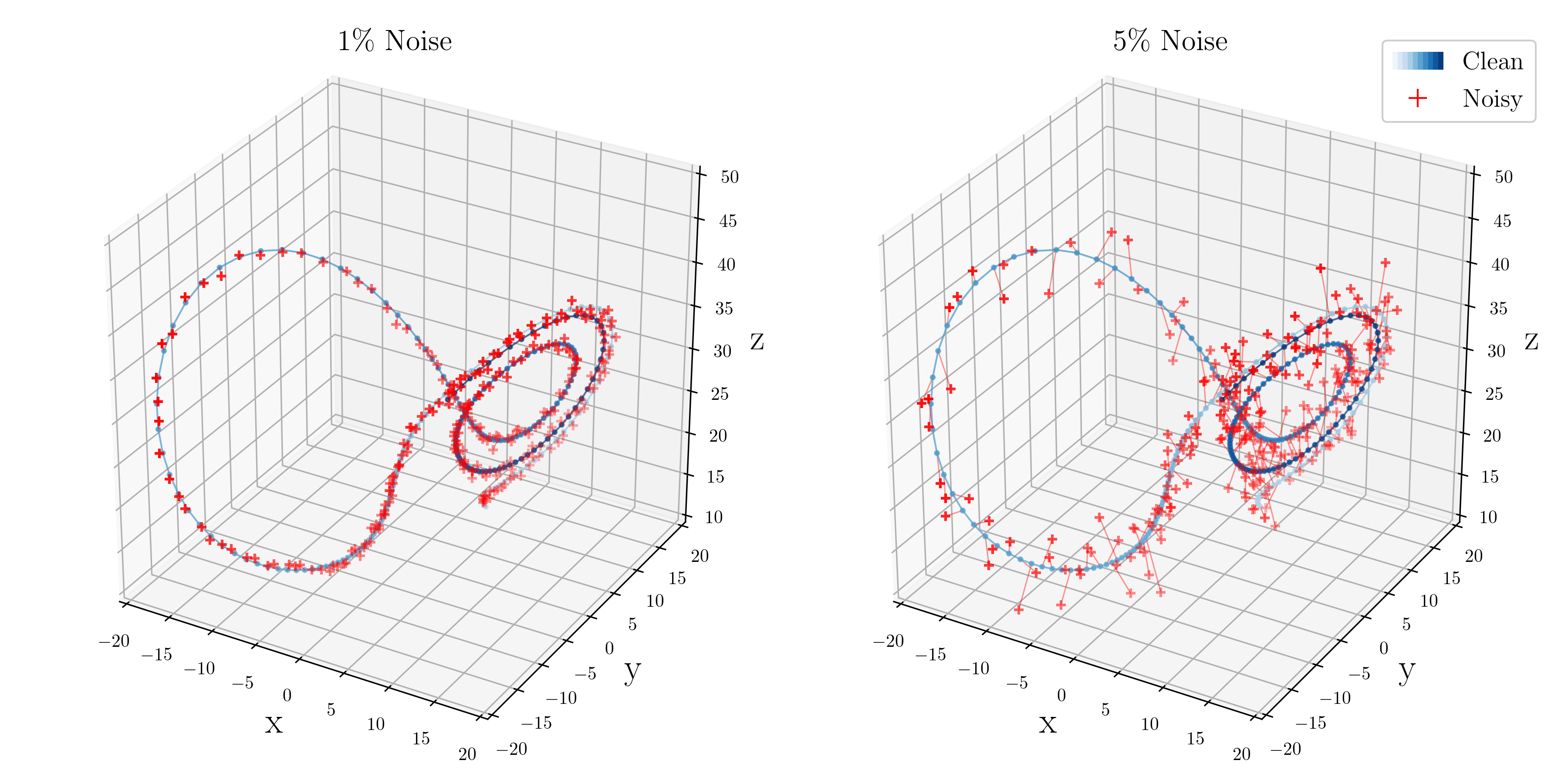}
    \caption{Comparison between the clean and contaminated (noisy) training data.}
    \label{fig:lorenz_model_noise}
\end{figure}

\noindent
\label{app:lorenz}
Predictions of the tested models for three test cases are plotted in Fig.~\ref{fig:lorenz_models_pred}.
All machine learning methods are accurate for the first several time-steps, but begin to quickly deviate from the numerical solution in subsequent time-steps. 
The transformer model is consistently accurate within the plotted time frame, typically outperforming alternative methods at later time-steps.
The predictions of the transformer model are also compared to the solution from a less accurate Euler time-integration method in Fig.~\ref{fig:lorenz_numeric_pred}.
Due to the numerical differences between the Runge-Kutta ground truth and the Euler methods, the solutions are quick to deviate and the transformer clearly outperforms the Euler numerical simulator.
Lastly, in Fig.~\ref{fig:lorenz_model_noise} we illustrate the two different noise levels used to contaminate the training data for the results listed in Table~\ref{table:lorenz_error_noise}.

\vspace{-2em}
\vspace{-2em}

\section{Cylinder Supplementary Results}
\begin{figure}[h]
    \centering
    \includegraphics[width=0.95\textwidth]{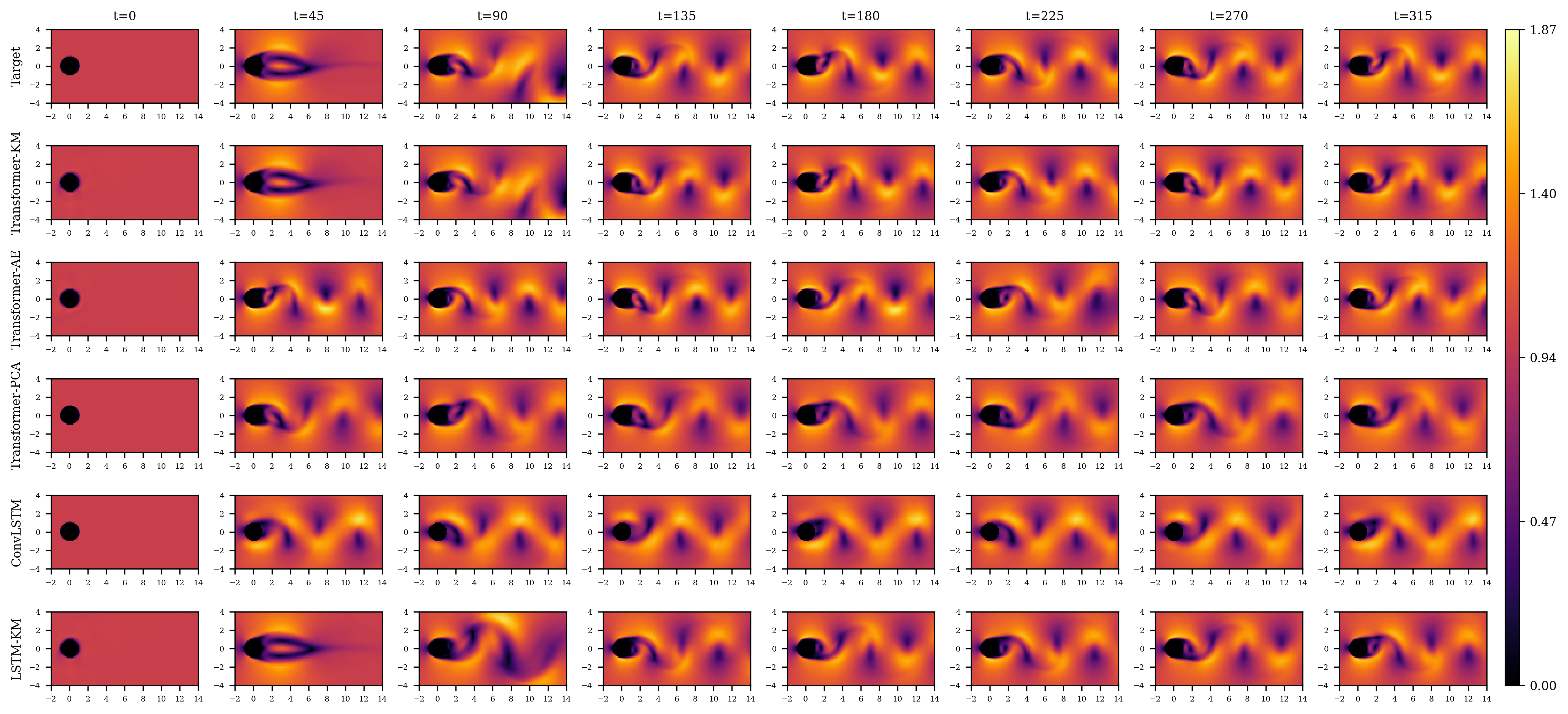}
    \caption{Velocity magnitude predictions of a test case at $Re=633$ using the transformer with Koopman (KM), auto-encoder (AE) and PCA embedding methods, the convolutional LSTM model (ConvLSTM) and fully-connected LSTM with Koopman embeddings (LSTM-KM).}
    \label{fig:cylinder_mag}
\end{figure}

\begin{figure}[h]
    \centering
    \includegraphics[width=0.95\textwidth]{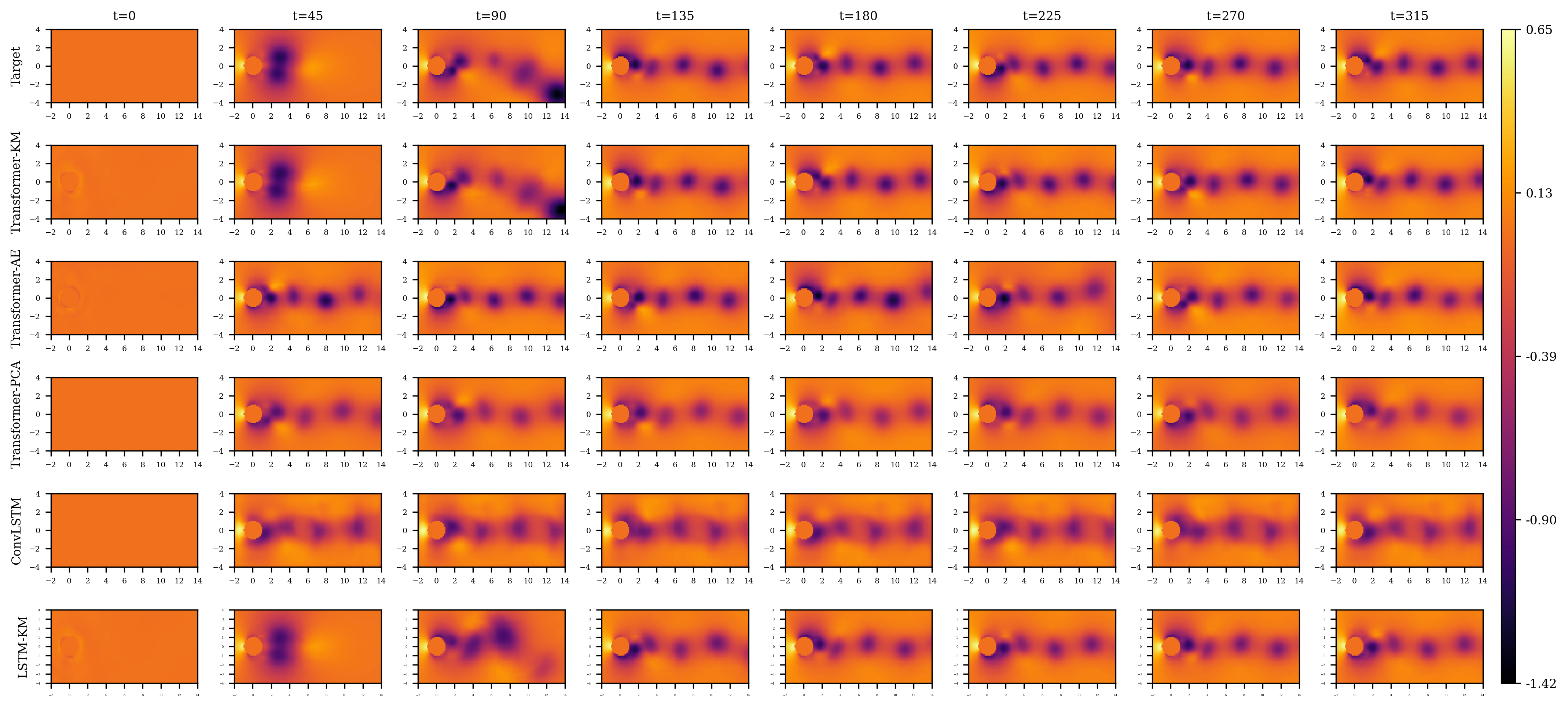}
    \caption{Pressure predictions of a test case at $Re=633$ using the transformer with Koopman (KM), auto-encoder (AE) and PCA embedding methods, the convolutional LSTM model (ConvLSTM) and fully-connected LSTM with Koopman embeddings (LSTM-KM).}
    \label{fig:cylinder_pressure}
\end{figure}

\begin{figure}[h]
    \centering
    \includegraphics[width=0.9\textwidth]{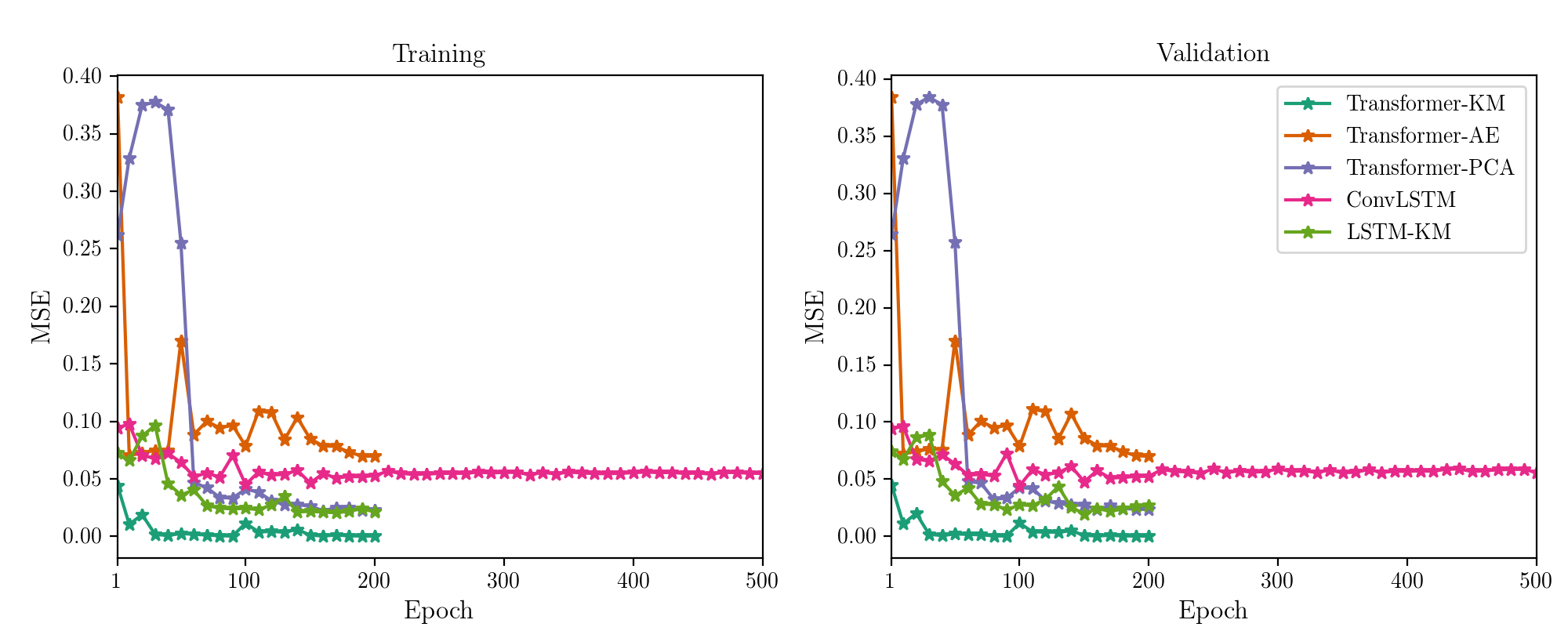}
    \caption{The training and validation mean square error (MSE) of the transformer with Koopman (KM), auto-encoder (AE) and PCA embedding methods, the convolutional LSTM model (ConvLSTM) and fully-connected LSTM with Koopman embeddings (LSTM-KM) during training.}
    \label{fig:cylinder_training}
\end{figure}

\noindent
\label{app:cylinder}
Prediction fields of all the tested models for a single test case are plotted in Figs.~\ref{fig:cylinder_mag} and~\ref{fig:cylinder_pressure}.
While all models are able to perform adequately with qualitatively good results, the transformer model with Koopman embeddings (Transformer-KM) outperforms alternatives.
Additionally, the training and validation errors during training are plotted in Fig.~\ref{fig:cylinder_training} where all models have similar training and validation error indicating minimal overfitting.

\begin{figure}[h]
    \centering
    \begin{subfigure}{\textwidth}
        \centering
        \includegraphics[width=\textwidth]{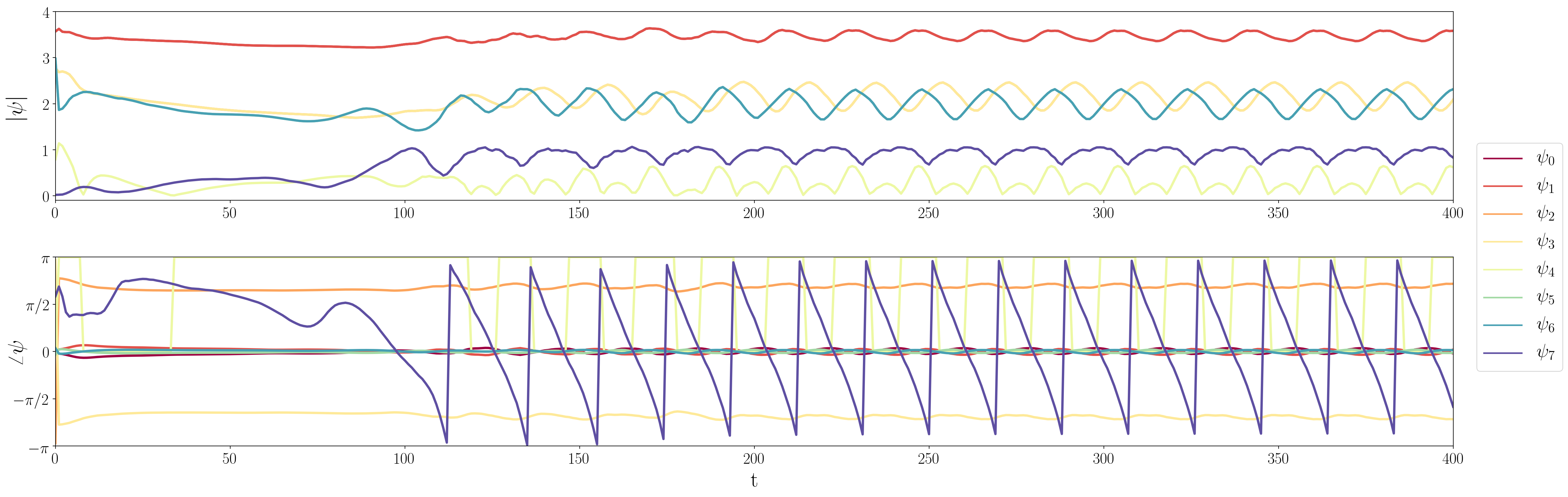}
        \caption{$Re=233$}
    \end{subfigure}
    \begin{subfigure}{\textwidth}
        \centering
        \includegraphics[width=\textwidth]{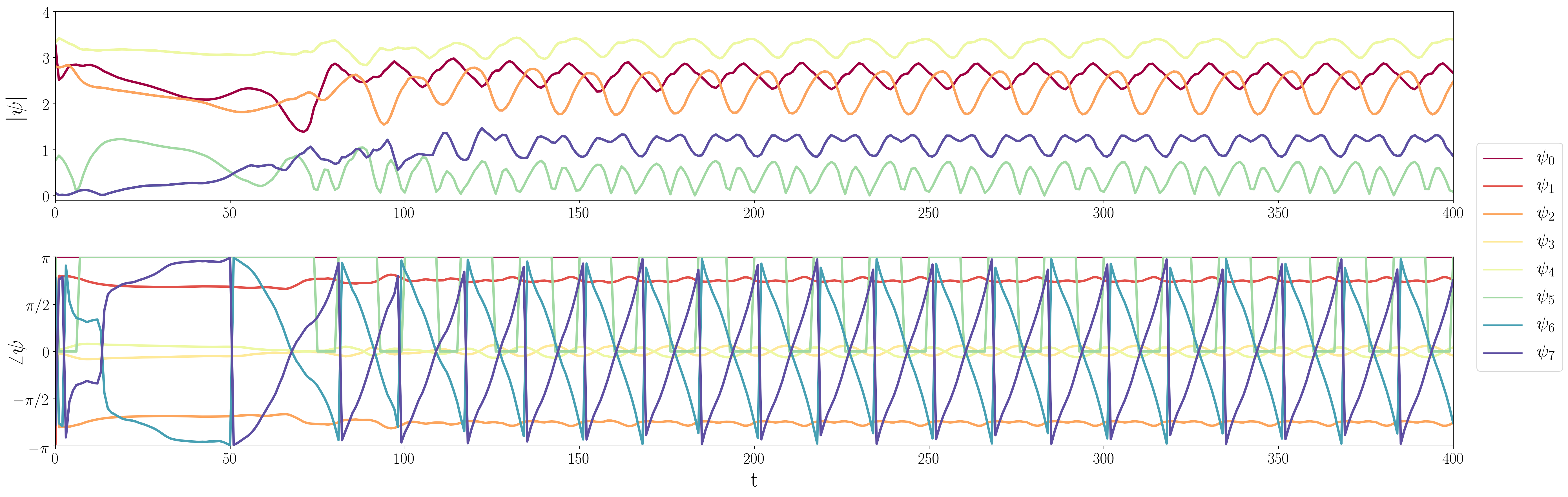}
        \caption{$Re=633$}
    \end{subfigure}
    \caption{The dynamics of the fluid flow around a cylinder projected onto the $8$ most dominant eigenvectors of the learned Koopman operator, $\mathcal{K}$, in the embedding model.}
    \label{fig:cylinder_koopman_modes}
\end{figure}

The evolution of the flow field projected onto the dominant  eigenvectors of the learned Koopman operator for two Reynolds numbers is plotted in Fig.~\ref{fig:cylinder_koopman_modes}.
This reflects the dynamical modes that were learned by the embedding model to impose physical ``context'' on the embeddings.
Given that the eigenvectors are complex, we plot both the magnitude, $|\psi|$, and angle, $\angle\psi$.
For both Reynolds numbers, it is easy to see the initial transition region, $t<100$, before the system enters the periodic vortex shedding.
Once in the periodic region, we can see that the higher Reynolds number has a higher frequency which reflects the increased vortex shedding speed.
\vspace{-2em}

\section{Gray-Scott Supplementary Results}
\begin{figure}[h]
    \centering
    \begin{subfigure}{\textwidth}
        \centering
        \includegraphics[width=0.9\textwidth]{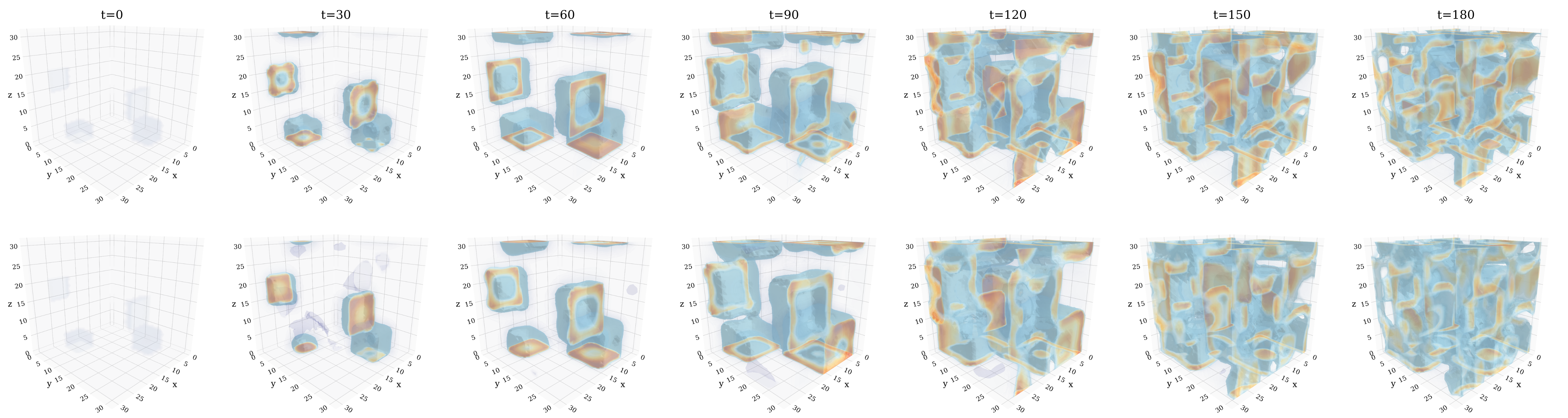}
        \caption{$\bm{u}$ target (top) and transformer prediction (bottom)}
    \end{subfigure}\\
    \begin{subfigure}{\textwidth}
        \centering
        \includegraphics[width=0.9\textwidth]{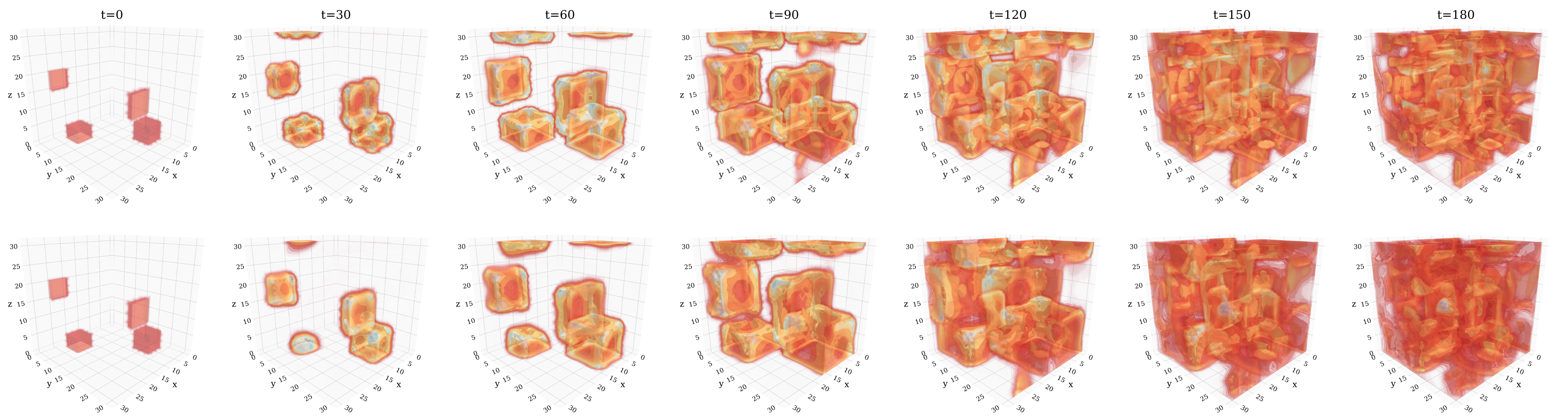}
        \caption{$\bm{v}$ target (top) and transformer prediction (bottom)}
    \end{subfigure}
    \caption{Test case volume plots for the Gray-Scott system. Isosurfaces displayed span the range $\bm{u},\bm{v}=[0.3,0.5]$ to show the inner structure.}
    \label{fig:gray_scott_app1}
\end{figure}

\begin{figure}[h]
    \centering
    \begin{subfigure}{\textwidth}
        \centering
        \includegraphics[width=0.9\textwidth]{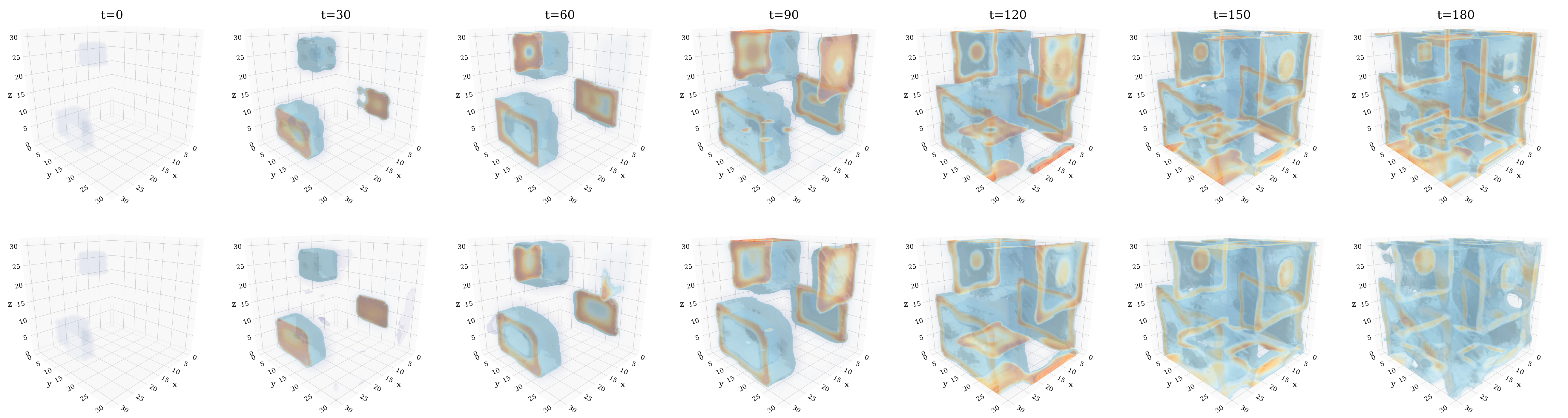}
        \caption{$\bm{u}$ target (top) and transformer prediction (bottom)}
    \end{subfigure}\\
    \begin{subfigure}{\textwidth}
        \centering
        \includegraphics[width=0.9\textwidth]{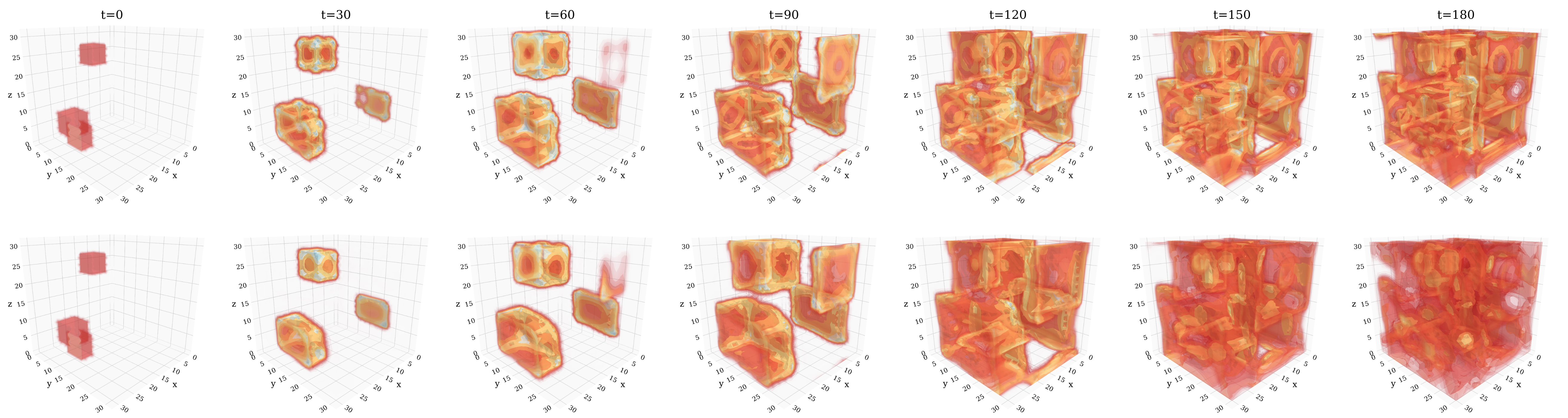}
        \caption{$\bm{v}$ target (top) and transformer prediction (bottom)}
    \end{subfigure}
    \caption{Test case volume plots for the Gray-Scott system. Isosurfaces displayed span the range $\bm{u},\bm{v}=[0.3,0.5]$ to show the inner structure.}
    \label{fig:gray_scott_app2}
\end{figure}

\begin{figure}[h]
    \centering
    \begin{subfigure}{\textwidth}
        \centering
        \includegraphics[width=\textwidth]{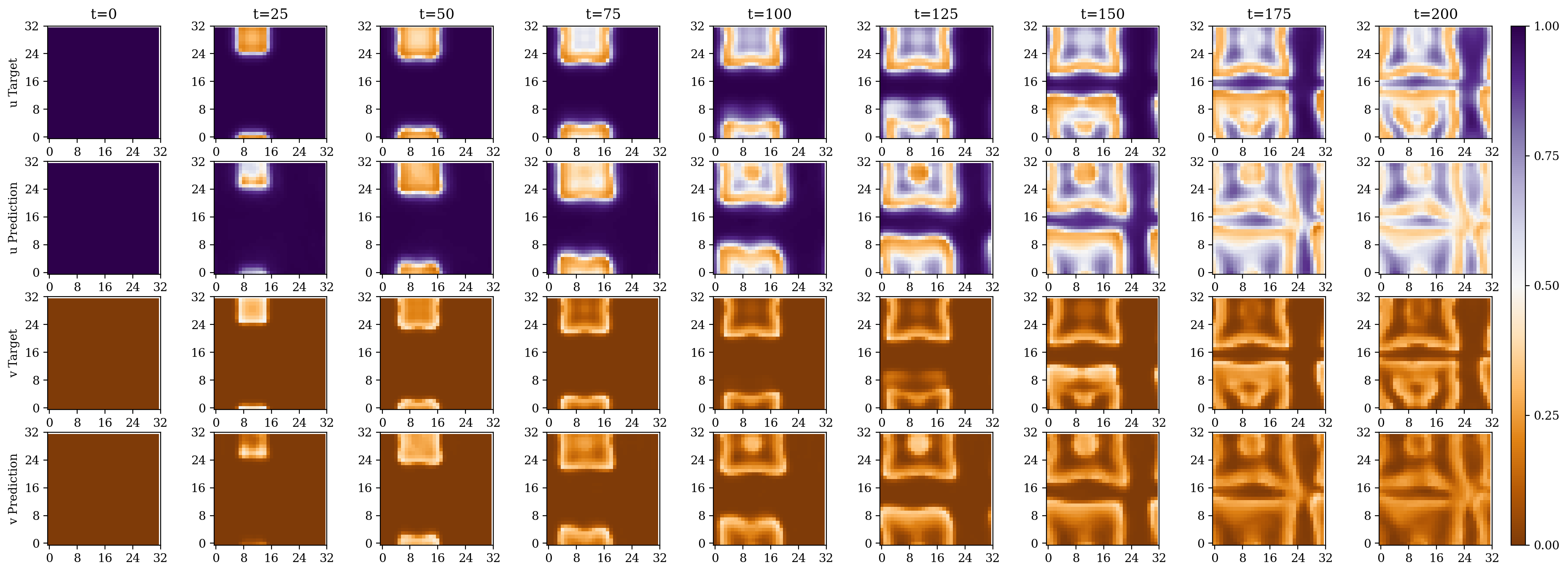}
    \end{subfigure}\\
    \vspace{1em}
    \begin{subfigure}{\textwidth}
        \centering
        \includegraphics[width=\textwidth]{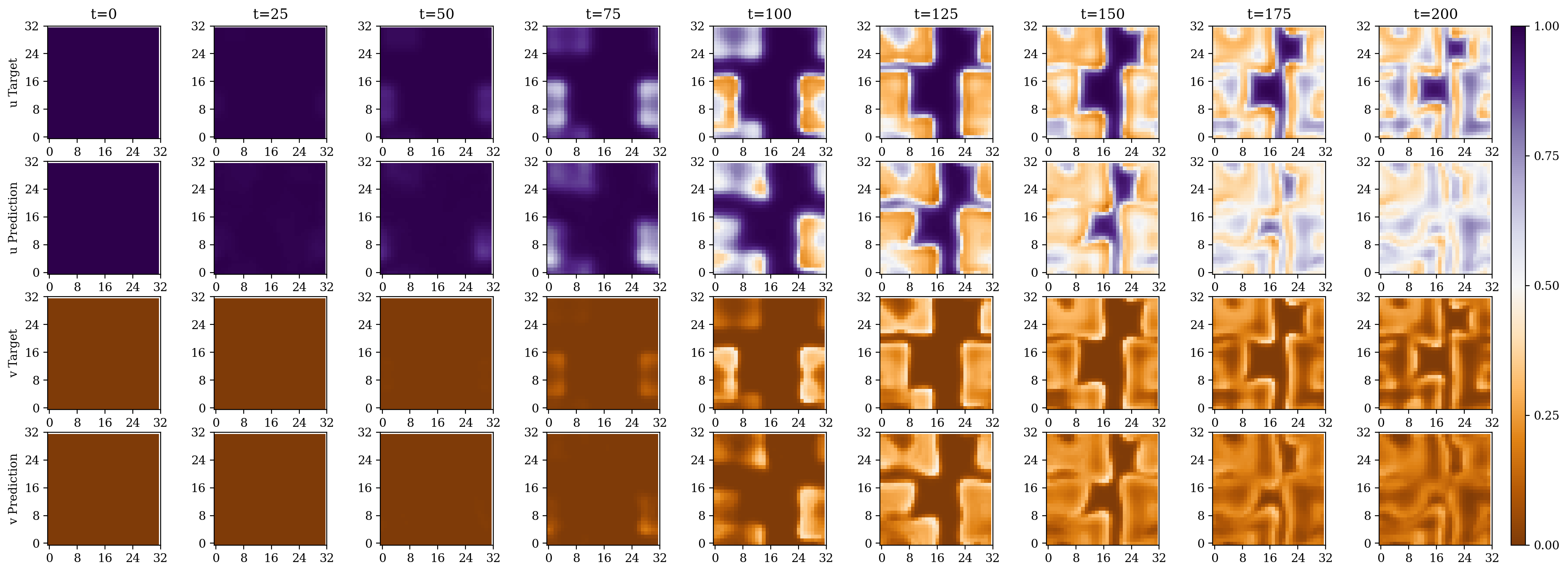}
    \end{subfigure}\\
    \vspace{1em}
    \begin{subfigure}{\textwidth}
        \centering
        \includegraphics[width=\textwidth]{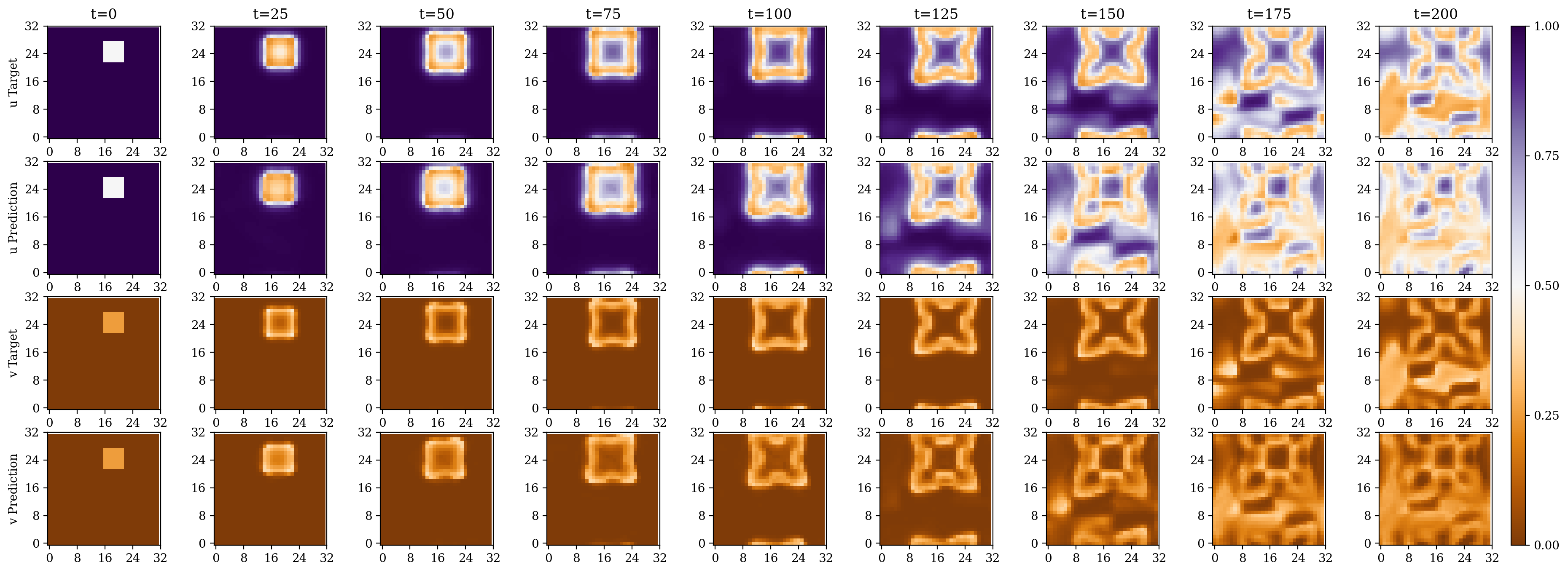}
    \end{subfigure}
    \caption{$x-y$ plane contour plots of three Gray-Scott test cases sliced at $z=16$.}
    \label{fig:gray_contours}
\end{figure}

\noindent
\label{app:gray_scott}
Volume plots of two test cases for both species are provided in Figs.~\ref{fig:gray_scott_app1} \&~\ref{fig:gray_scott_app2}.
In addition to better visualize the accuracy of the trained transformer model, contour plots for three test cases are also provided in Fig.~\ref{fig:gray_contours}.
The transformer model is able to reliably predict the earlier time-steps with reasonable accuracy.
As the system continues to react, the reaction fronts begin to interact resulting in the well-known complex structures of the Gray-Scott system~\cite{pearson1993complex, lee1993pattern}.
During these later times, the transformer begins to degrade in accuracy yet does maintain notable consistency with the true solution.

\end{document}